\documentclass{article}

\usepackage[preprint]{neurips_2025}

\usepackage[utf8]{inputenc} 
\usepackage[T1]{fontenc}    
\usepackage{hyperref}       
\usepackage{url}            
\usepackage{booktabs}       
\usepackage[table]{xcolor}
\usepackage{amsfonts}       
\usepackage{amsmath}
\usepackage{amssymb}
\usepackage{mathtools}
\usepackage{amsthm}
\usepackage{nicefrac}       
\usepackage{microtype}      
\usepackage{xcolor}         
\usepackage{enumitem}
\usepackage[textsize=tiny]{todonotes}
\usepackage{listings}
\usepackage{pdflscape}
\usepackage{pgf}
\usepackage{pgfplots}
\usepackage{multirow}
\usepackage{wrapfig}
\usepackage[capitalize,noabbrev]{cleveref}

\lstdefinestyle{yaml}{
    numbers=left,
    basicstyle=\ttfamily\footnotesize,
    emph={name,datasets,transformations,embedders,measures,seeds,data_feeds,use_cache,workers,use_database,rebuild_image,record_runtime,restart_failed,test_time_limit,compare_results_to},
    emphstyle=\color{orange},
    numberstyle=\tiny\color{gray},
    rulecolor=\color{orange},
    string=[s]{'}{'},
    stringstyle=\color{blue},
    comment=[l]{:},
    commentstyle=\color{black},
    morecomment=[l]{-},
    xleftmargin=17pt
 }
 
\usetikzlibrary{positioning,calc,decorations}
\usepgfplotslibrary{polar}
\pgfplotsset{compat=newest}
\pgfmathsetseed{42}
\tikzset{cross/.style={cross out, draw, 
         minimum size=2*(#1-\pgflinewidth), 
         inner sep=0pt, outer sep=0pt}}

\pgfmathdeclarefunction{gauss}{2}{%
  \pgfmathparse{1/(#2*sqrt(2*pi))*exp(-((x-#1)^2)/(2*#2^2))}%
}
\newcommand{\intersect}{\lceil \frac{k}{2} \rceil}
\newcommand{\mcrot}[1]{\rlap{\rotatebox{60}{#1}~}} 

\newcommand*{\twoelementtable}[3][l]%
{%  
    \begin{tabular}[t]{@{}#1@{}}%
        #2\tabularnewline
        #3%
    \end{tabular}%
}

\title{STEB: In Search of the Best Evaluation Approach for Synthetic Time Series}

\author{%
  Michael Stenger\\ 
  University of Wuerzburg\\
  97074 Wuerzburg, Germany \\
  \texttt{michael.stenger@uni-wuerzburg.de}\\
  \And
  Robert Leppich\\
  University of Wuerzburg\\
  97074 Wuerzburg, Germany\\
  \texttt{robert.leppich@uni-wuerzburg.de}\\
  \AND
  Andr\'e Bauer\\
  Illinois Institute of Technology\\ 
  Chicago, US\\
  \texttt{abauer7@iit.edu}\\
  \And
  Samuel Kounev\\
  University of Wuerzburg\\
  97074 Wuerzburg, Germany\\
  \texttt{samuel.kounev@uni-wuerzburg.de}\\
}

\newcommand{\acs}{ACS}
\newcommand{\apen}{ApEn} 
\newcommand{\authenticity}{Authenticity}
\newcommand{\autocorr}{Autocorrelation}
\newcommand{\ctwost}{C2ST}

\newcommand{\cas}{CAS}
\newcommand{\contextfid}{Context-FID}
\newcommand{\coverage}{Coverage}
\newcommand{\density}{Density}
\newcommand{\detectionmlp}{Detection\_MLP}
\newcommand{\detectionxgb}{Detection\_XGB}
\newcommand{\detectiongmm}{Detection\_GMM}
\newcommand{\detectionlinear}{Detection\_linear}
\newcommand{\discriminative}{Discriminative score}
\newcommand{\distributionalmetric}{Distributional metric}
\newcommand{\domias}{DOMIAS}
\newcommand{\fbca}{FBCA}
\newcommand{\icd}{ICD} 
\newcommand{\improvedprecision}{Improved precision}
\newcommand{\improvedrecall}{Improved recall}
\newcommand{\innd}{INND}
\newcommand{\jsd}{JSD}
\newcommand{\kld}{KLD}
\newcommand{\maxrts}{Max-RTS}
\newcommand{\mtopdiv}{MTop-Div}
\newcommand{\ndb}{NDB}
\newcommand{\ndbou}{NDB-over/under}
\newcommand{\onnd}{ONND}
\newcommand{\predictive}{Predictive score}
\newcommand{\rts}{RTS}
\newcommand{\sigmmd}{Sig-MMD}
\newcommand{\spatial}{Spatial correlation}
\newcommand{\sts}{STS}
\newcommand{\temporal}{Temporal correlation}
\newcommand{\trts}{TRTS}
\newcommand{\tstr}{TSTR}
\newcommand{\wcs}{Wavelet coherence score}
\newcommand{\wsd}{WD}
\newcommand{\alphaprecision}{$\alpha$-Precision}
\newcommand{\betarecall}{$\beta$-Recall}

\newcommand{\distvisual}{Distribution visualization}
\newcommand{\realism}{Realism score}
\newcommand{\visual}{Visual assessment}

\begin{document}

\maketitle

\begin{abstract}
The growing need for synthetic time series, due to data augmentation or privacy regulations, has led to numerous generative models, frameworks, and evaluation measures alike. Objectively comparing these measures on a large scale remains an open challenge. We propose the \textbf{S}ynthetic \textbf{T}ime series \textbf{E}valuation \textbf{B}enchmark (STEB)---the first benchmark framework that enables comprehensive and interpretable automated comparisons of synthetic time series evaluation measures. Using 10 diverse datasets, randomness injection, and 13 configurable data transformations, STEB computes indicators for measure reliability and score consistency. It tracks running time, test errors, and features sequential and parallel modes of operation. In our experiments, we determine a ranking of 41 measures from literature and confirm that the choice of upstream time series embedding heavily impacts the final score.
\end{abstract}

\section{Introduction}

Time series (TS) data is central to many domains and applications such as forecasting medical data~\cite{kaushik2020ai} or human activity recognition~\cite{gu2022survey}. Yet too often there is a lack of availability, quantity, and quality of data, for instance, due to privacy concerns. Data synthesis can be a solution, but it remains challenging in practice~\cite{fonseca2023tabular, Brophy2023Generative}. A key step towards high-quality synthetic data is a comprehensive, reliable evaluation strategy. The fundamental problem is the lack of ground truth data similar to unsupervised clustering. To handle this complex, indirect assessment, the common approach is to use different measures to quantify various quality aspects~\cite{time-gan}. However, with dozens of measures having been proposed in recent years, the selection of the ``ideal'' set of measures is particularly challenging~\cite{stenger2024evaluation}. For instance, seminal works~\cite{time-gan,xu2020cot,coscigan} and recently proposed frameworks~\cite{synthcity, tsgm} use different combinations with little overlap. Furthermore, the aspects of synthetic data quality considered typically differ~\cite{stenger2024categories}. As a consequence, comparability of generative performance is hindered, the state-of-the-art unknown. Previous works studied small groups of measures in a tailored analysis or developed frameworks to generate and evaluate synthetic data. However, to the best of our knowledge, there is no detailed, objective, and broad study on the effectiveness of these measures applied to TS. Critically, this raises serious concerns regarding the reliability of past evaluation results for TS generative models. Hence, we propose \textbf{S}ynthetic \textbf{T}ime series \textbf{E}valuation \textbf{B}enchmark~(STEB), the first benchmark framework to conduct large-scale and multi-faceted analysis of synthetic TS evaluation measures. With STEB, we aim to narrow down the clutter of measures to a standardized set, drastically increasing the comparability of generative performance.

\textbf{Scope.} We focus on uni-/multivariate, real-valued TS. An evaluation measure is a function $m: \mathcal{P}(\mathcal{X}) \times \mathcal{P}(\mathcal{X}) \rightarrow \mathbb{R}$ where $\mathcal{P}$ is the super-set operator and $\mathcal{X}$ the data space, e.g., time series. For a real dataset $D_r \subset \mathcal{X}$ and synthetic dataset $D_s \subset \mathcal{X}$, we call $s = m(D_r, D_s)$ its score. In this context, ``real'' data is the initial data to learn. Note that $m$ assesses the generated data, not generators. Hence, we do not consider generator-dependent measures such as duality gap~\cite{duality-gap}. Similarly, we limit this study to quantitative measures with clear $s$ and exclude, for instance, visualizations such as the popular t-SNE plot~\cite{time-gan}.

\textbf{Contributions.} With this work, we contribute to synthetic data research in three ways:
\vspace{-3mm}
\begin{enumerate}[itemsep=0pt]
    \item We design and implement STEB, a novel benchmark framework for comprehensive, interpretable, and automated analysis of quantitative synthetic TS measures. 
    \item We analyze 41 measures with respect to reliability, consistency, and running time. More specifically, we rank the measures in four aspects of quality and in running time.
    \item We investigate the impact of the upstream TS embedding on the final score.
\end{enumerate}
\vspace{-2mm}

This paper is structured as follows. \cref{sec:related_work} puts our work in the context of related work. \cref{sec:steb} introduces STEB and \cref{sec:experiments} the experiments on TS synthesis measures. \cref{sec:results} presents and discusses the results. \cref{sec:conclusion} concludes the paper.
\section{Related Work} 
\label{sec:related_work}

With the increasing interest in synthetic data generation, different studies on evaluation measures were conducted. One noteworthy work examines three of the then most commonly used evaluation measures for image generation~\cite{theis2016note}. Their combined theoretical and empirical approach is tailored towards each measure. Key findings are that the behavior of different measures is often largely independent of each other and synthetic data utility is application specific. Lucic et al.~\cite{lucic2018are} presented an empirical study on GAN models and evaluation measures for image synthesis. Their focus is on the Fréchet inception distance (FID), analyzing bias, variance, robustness to mode dropping, different image embeddings, and FID value range. In terms of experimental design, the closest work we know of is by Huang et al.~\cite{huang2018an}. They manipulate image data in different ways to study the behavior and efficiency of five evaluation measures w.r.t. overfitting, mode collapse/dropping, discriminability, and robustness.  Recently, Ismail-Fawaz et al.~\cite{ismail2024establishing} conducted a comparison of eight measures for the evaluation of human motion generation and proposed a ninth. They analyzed each measure qualitatively, followed by a quantitative comparison using a conditional generator on one real dataset to controllably create human motions.

STEB differs from the above works in many ways: (i) While these work are all (very) focused in their analysis, we compare a wide range of measures of diverse designs and purposes; (ii) STEB is designed for TS, whereas previous works target images and human motion; (iii) STEB is model agnostic with regard to the data generator and evaluation measure, unlike Lucic et al.~\cite{lucic2018are} and Huang et al.~\cite{huang2018an}, who focus on non-conditional GANs, or Theis et al.~\cite{theis2016note}, who employ analysis specific to the measures; (iv) We utilize different hand-crafted data manipulations, while Ismail-Fawaz et al.~\cite{ismail2024establishing} use a neural model to create the test synthetic data; (v) Most of the related work does not capture recent developments. STEB incorporates established and recent  measures and  can be extended to support further analysis techniques and future measures, continuously tracking the state-of-the-art; (vi) STEB is more fine-grained in its analysis, as it differentiates four aspects of synthesis quality, and it is more comprehensive in terms of experimental parameters.

There are three related synthetic data benchmarks. Synthcity is a framework for benchmarking tabular, image, and TS data generators~\cite{synthcity}. It incorporates multiple generators, evaluation measures, and datasets in an automated test pipeline. Similarly, TSGBench~\cite{tsgbench} and Time Series Generative Modeling (TSGM)~\cite{tsgm} are frameworks for time series synthesis. While TSGBench specializes on benchmarking generators, TSGM is presented as a more general solution including application cases. These benchmarks mainly differ in the choice of integrated measures, generators, and datasets. They evaluate the respective generative models and accept the implemented measures as given. STEB, however, assesses the measures themselves in order to determine which to best include in tools such as Synthcity, TSGBench, or TSGM. The diversity of evaluation suites highlights the need for a systematic analysis tool for measures. 

\section{STEB: Synthetic Time Series Evaluation Benchmark}
\label{sec:steb}

In this section, we introduce the approach and present the design of STEB, the first benchmark framework for analyzing evaluation measures for synthetic TS. 

\subsection{Controlled Distribution Modulation}
\label{subsec:steb:approach}

The evaluation of synthetic data $D_s$ is commonly centered around given data generators with unknown performance on a set of real data $D_r$. The absence of ground truth complicates analysis and comparison, leading to the development of complex evaluation measures. To this end, we change the perspective and select benchmark scenarios with an expected outcome to evaluate the performance of the measures themselves. Inspired by Lucic et al.~\cite{lucic2018are}, we construct such scenarios by replacing the ``regular'' TS generator $G$ with a pseudo-generation method: \textit{Transformation}~$T$. Formally, $T: \mathbb{R}^{n \times l \times d} \times [0,1] \rightarrow \mathbb{R}^{n \times l \times d}$ is a function taking a dataset of $n$ real-valued TS of length $l$ and dimension $d$, along with a scale factor $\kappa$ specifying the transformation intensity. Its output is a ``transformed'' dataset $D_T^\kappa$. However, finding $T$ such that the score $s = m(D_r, D_T^\kappa)$ can be assessed directly and absolutely is challenging for the non-trivial case $\kappa > 0$. $D_T^\kappa$ must be complex enough to be a realistic test case, but determining the expected $s$ must be known for non-trivial input data. Additionally, different measures produce scores in varying ranges and optimization directions. 

\begin{figure}[t]
    \centering
    {\small
\begin{tikzpicture}
    \begin{axis}[
    no markers, domain=0:10.5, samples=100,
    axis lines*=left,
    height=38mm, width=98mm,
    xtick=\empty, ytick=\empty,
    enlargelimits=false, clip=false, axis on top
    ]
        \addplot [fill=cyan!20, draw=none, domain=0:3.25] {gauss(3.7,1.3)} \closedcycle;
        \addplot [fill=cyan!20, draw=none, domain=3.25:7] {gauss(2.5,1)} \closedcycle;
        \addplot [fill=red!20, draw=none, domain=0:4.5] {gauss(7.5,1.7)} \closedcycle;
        \addplot [fill=red!20, draw=none, domain=4.5:8] {gauss(2.5,1)} \closedcycle;
        \addplot [very thick,black] {gauss(2.5,1)};
        \addplot [very thick,cyan,dashed] {gauss(3.7,1.3)};
        \addplot [very thick,red,dotted] {gauss(7.5,1.7)};
        \node [yshift=-0.3cm] at (axis cs:2.5,0) {$P=P_T^0$};
        \node [yshift=-0.3cm, cyan] at (axis cs:3.7,0) {$P_T^1$};
        \node [yshift=-0.3cm, red] at (axis cs:7.5,0) {$P_T^2$};
    \end{axis}
    \draw[->] (2.8,2.4) -- (6.6,2.4);
    \node[fill=white] at (4.7, 2.4) {Modulation};
\end{tikzpicture}
}
    \caption{Depiction of the modulation concept. By modulating parameter $\kappa$, we can influence the degree to which a transformation $T$ impacts dataset $D_r$ (resp. its underlying distribution $P$) to create the pseudo-synthetic dataset $D_T$ with distribution $P_T$. For $\kappa = 0$, we get $P_T^0 = P$ (black), for $\kappa = 0.3$, it might be $P_T^{1}$ (blue, dashed), and for $\kappa = 0.9$, it is $P_T^{2}$ (red, dotted).}
    \label{fig:modulation}
\end{figure}
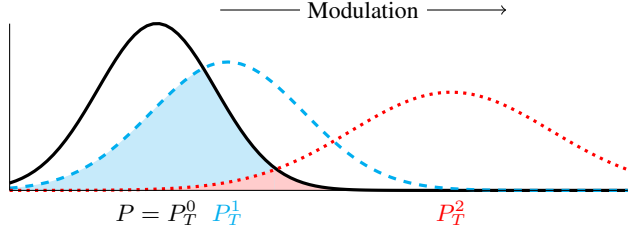

To address this issue, we combine transformations with a second concept, the \textit{modulation} of $T$ via intensity $\kappa$. Intuitively, we modify $D_r$ more and more, creating a series of ever more different synthetic data, and assess the score of each step relative to the others. More formally: 
$D_r$ is a sample drawn from an underlying distribution $P$ in $\mathbb{R}^{l \times d}$, and synthesizing $D_s$ equates to generating new samples from $P$. A $1$D simplification of $P$ and the modulation process is depicted in \cref{fig:modulation}. $\kappa$ allows us to create a sample $D_T^\kappa$ of a shifted and distorted distribution $P_T$ in the data space, ranging from $P$ itself to a completely different distribution. Assuming $m$ measures some aspect of similarity of the underlying distributions of two datasets, we expect $s$ to get worse with increasing $\kappa$. This is the expected outcome we can test empirically. More specifically, we test if $s_0 > s_1 > s_2 > \dots$ (assumption: higher is better) for $s_i = m(D_r, D_T^{\kappa_i})$ along the ``modulation path'' $\kappa_0 < \kappa_1 < \kappa_2 < \dots$. Using this condition, we compute a reliability indicator for $m$ under different test cases, varying $T$, $D_r$, and the random seed. The average value across all tests serves as approximation for the measure's reliability.

\textbf{Example.} Let 
\vspace{-4pt}
\begin{equation}
    m_{\text{iMAE}}: D \times D' \mapsto \left(10^{-3} + \frac{1}{nld} \sum_{i, j, k} |D_{i,j,k} - D'_{i,j,k}| \right)^{-1}   
\end{equation}
 be the inverse mean absolute error on ordered $D, D'$. Further, let $T: D, \kappa \mapsto \{ x+\kappa \mid x \in D \}$ be a transformation increasing every scalar in every TS in D by the scale factor. We compute $D_T^0 = T(D, 0)$, $D_T^{0.5} = T(D, 0.5)$, $D_T^1 = T(D, 1)$ and $s_0 = m_\text{iMAE}(D, D_T^0)$, $s_1 = m_\text{iMAE}(D, D_T^{0.5})$, $s_2 = m_\text{iMAE}(D, D_T^1)$. As $m_\text{iMAE}$ measures the inverse average distance between the scalars of two datasets and $T$ increases these values with increasing $\kappa$, we find that $s_0 > s_1 > s_2$. Hence, $m_\text{iMAE}$ behaves as expected and we would assign a high reliability indicator. Note that the regular MAE would satisfy none of these inequalities, resulting in bad performance.
    
\subsection{Transformations}

In the following, we list all 13 transformations implemented in STEB and used in our experiments, chosen based on the diversity of data changes they force the measures to detect, their ability to allow gradual transformation with $\kappa$, a sensible running time, and the interpretability of induced changes. 

\textbf{Gaussian noise} adds a matrix of random values sampled from a Gaussian distribution with 0 mean and $\frac{\kappa}{2}$ variance to the TS in $D_r$. To standardize the amount of noise for each dataset, we scale $D_r$ to $[0,1]$ before applying the noise and rescale it afterwards.

\textbf{Label corruption} corrupts the labeling of classified datasets by randomly swapping the labels of $\frac{\kappa}{10}$ of instances. Only applicable to measures sensitive to labels.

\textbf{Misalignment} rotates the different channels of each TS in $D_r$ randomly by $p$ positions with probability $\kappa$, $ 1 \le p \le \kappa(l-1)$, and narrows the gap between the formerly first and last values. Only applicable to multivariate TS.

\textbf{Mode collapse} simulates a mode collapse by sampling each class of $D_r$ down by $\kappa$, replacing the dropped instances in each mode with noisy duplicates of the remaining time series. Only applicable to labeled data.

\textbf{Mode dropping} simulates mode dropping by replacing all TS of $\kappa$ classes in $D_r$ with TS from the remaining classes proportional to their size. Only applicable to labeled data.

\textbf{Moving average} transforms $D_r$ by applying a moving average to each channel of each TS. The filter used for averaging is $a\cdot l\cdot \kappa + 1$ wide where $a=\frac{1}{3}$ if $l \ge 30$ else $a=1$ and centered on the modified value.

\textbf{Rare event drop} probes the sensitivity to rare events, in this case, the smallest class in $D_r$. $D_T$ is created by swapping out $\kappa$ of its instances with ones of other classes taken from another substitute dataset $D_\text{rs}$. This set provides additional real TS from the same distribution exclusively accessed by transformations. Only applicable to labeled data.

\textbf{Reverse substitution} probes the sensitivity to leaking real TS into the synthetic set. It starts with $D_\text{rs}$ and gradually adds up to ten $D_r$ instances to $D_T$ with increasing $\kappa$.

\textbf{Salt \& pepper} adds noise to the data by replacing random values in $D_r$ by 0 and 1 with probability $\frac{\kappa}{2}$.

\textbf{Segment leaking} builds the output $D_T$ by using $D_\text{rs}$ as a basis and replacing $30\kappa$ random segments from TS in $D_\text{rs}$ with segments from $D_r$. A segment is one channel of a TS subsequence and between $\frac{l}{4}$ and $\frac{l}{2}$ long.

\textbf{STL decomposition} transforms $D_r$ by decomposing every channel $c$ of every TS into season $s$, trend $t$, and residual $r$ using LOESS~\cite{cleveland1990stl} followed by its reconstruction via linear combination $c_\text{new} = (\kappa u_1 + 1) s + (\kappa u_2 + 1) t + (\kappa u_3 + 1) r)$ with $u_1, u_2, u_3 \sim \mathcal{U}_{[-1,1]}$.

\textbf{Substitution} replaces a fraction $\kappa$ of the TS in $D_r$ with instances from $D_\text{rs}$ at random.

\textbf{Wavelet transform} decomposes each TS in $D_r$ with discrete wavelet transform~\cite{lee2019pywavelets}, rescales its scale component by $\kappa$, and inverses the transformation again. This gradually increases/removes the temporal structure of each channel while leaving the residuals intact.

Note that none of the presented transformations is expected to mimic synthetic time series produced by typical generative models, which are practically intractable. Instead, they are intended to check individual, tractable aspects of relevant measure behavior. Collectively, the transformations cover a wide and diverse range of measure behavior across hundreds of tests. This is comparable to classification or forecasting, where the evaluation with fixed settings and dataset selection serves as approximation of general model performance.

\subsection{Benchmark Design}
\label{subsec:steb:design}

\begin{wrapfigure}{R}{0.5\textwidth}
    \includegraphics[width=\linewidth]{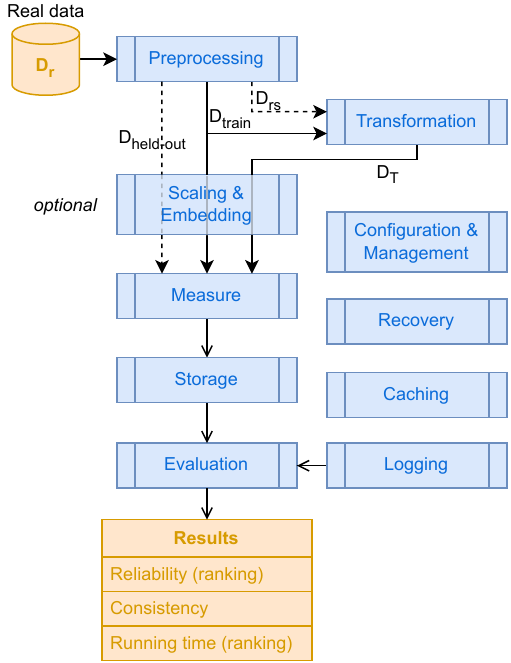}
    \caption{Architectural design of STEB. Input (top left) and output (bottom) are highlighted in orange, STEB components in blue. Datasets are referenced as $D$ and the flow of data is indicated by arrows with filled head. The open-headed arrows mark information flow such as scores, rankings, measurements, and error messages. Dashed arrows denote conditional data flow, where $D_\text{held\_out}$ depends on the measure and $D_\text{rs}$ on the transformation.}
    \label{fig:design}
    \vspace{-5mm}
\end{wrapfigure}

The purpose of our benchmark is to enable the comprehensive, interpretable, and automated comparison of TS evaluation measures. Its main design goals are extensible components, fast and resource-efficient execution, and flexibility in operation. To achieve this, STEB is based on the distribution modulation and transformation concepts, while being centered around the measures under test and being supported by several auxiliary components. The high-level architecture and data-/information-flow is depicted in \cref{fig:design}. Each component is described in the next section.

STEB works with two organizational units, \textit{experiments} and \textit{tests}. An experiment represents one run of the benchmark; it is initiated by the user and specified through a configuration (e.g., see Listing~\ref{lst:config}). A test is one pass through STEB from the \emph{Transformation} component to \emph{Storage}; it is characterized by a set of parameters including the input dataset, the measure to test, and the transformation to apply. An experiment includes multiple, often thousands of tests. It starts with an initialization via the \emph{Configuration~\& Management} component, followed by the \emph{Preprocessing} of all required real data. Afterwards, the included tests are gradually processed, starting with the transformation to create $D_T$. If required by the measure, the test datasets are scaled to $[0, 1]$ or embedded. Finally, the data flow arrives at the \emph{Measure} component, where a score for $D_\text{train}$, $D_\text{T}$, and $D_\text{held-out}$ is computed. The transformation, scaling/embedding, and measure steps are repeated for each $\kappa$. Upon completion, the different scores are collected and stored, before a new test is selected by the \emph{Management} component. If a test incurs an exception, it will be recorded as failure, its reason logged, and the experiment resumed with the option to repeat the test at a later time. The \emph{Evaluation} component is called once all tests are processed. Measures, embedders, transformations, and datasets are integrated via common interfaces, making these components extensible.

\subsection{Components}
\label{subsec:steb:components}

Core STEB components are described below, supporting components for storage, caching, recovery, and logging are detailed in \cref{sec:appendix:benchmark}. 

\textbf{Data Preprocessing.} STEB implements a diverse group of ten datasets from different domains. A detailed list with description, source, and characteristics can be found in \cref{subsec:appendix:datasets}. We implemented an automated preprocessing pipeline to prepare each dataset in the specific format required by each measure and transformation. It retrieves the specified data from its online source, removes outliers, interpolates missing values, equalizes TS lengths, extracts class labels, and calculates dataset statistics. Depending on the input requirements of transformation and measure, the real data is split into up to three, usually two, equally sized subsets $D_\text{train}$, $D_\text{rs}$, and $D_\text{held-out}$. $D_\text{train}$ represents the data available to a potential generator, $D_\text{rs}$ is the substitute data, and $D_\text{held-out}$ simulates generator test data. The latter two are optional. 

\textbf{Scaling \& Embedding.} Many measures require further preprocessing of the input data. One group works well only on scaled values, another (overlapping) group operates on real-valued vectors and is not directly applicable to TS data (especially if they are multivariate). The scaling to range $[0,1]$ is straightforward. To facilitate vector representations, this component has three embedders implemented. By default, the \textit{TS2Vec}~\cite{ts2vec} representation model is employed. Alternatively, it features a non-DL-based embedder called \textit{Catch22}~\cite{lubba2019catch22} and the trivial concatenation of feature channels referred to as \textit{Concat} $e_\text{concat}: \mathbb{R}^{l \times d} \rightarrow \mathbb{R}^{l \cdot d}$ as baseline model.

\textbf{Measure.} STEB includes 44 measures collected over the past years from different backgrounds with diverse evaluation goals and varying levels of complexity. Two are qualitative, visualization-based measures, while the others are quantitative measures. They usually produce a score $s \in \mathbb{R}^+_0$, but there are outliers such as C2ST~\cite{two-sample} with $s \in \{ \texttt{true}, \texttt{false}\}$. Moreover, the measures differ in terms of input data. Three measures only consider synthetic data, here provided as $D_T$, while most~(33) also consider the real data provided to the generator, $D_\text{train}$. The remaining eight require a third, real dataset unknown to the generator, $D_\text{held-out}$. Due to the space limitations, we prepared an alphabetical list of the 44 measures with descriptions and sources in \cref{sec:appendix:measures}. For implementation details, we refer to the STEB code and documentation.

\textbf{Evaluation.} Once all tests in an experiment are concluded, whether successful or not, the evaluation can start. To this end, the scores produced for the tested measures and the recorded running times (if available) are analyzed by this component based on three criteria: (i)~the \textit{reliability} of a measure to truthfully and accurately reflect the quality of a given synthetic dataset in its score; (ii)~the \textit{consistency} of the measure's scores with respect to changing parameters such as random seed or dataset; (iii)~the measure's speed in computing a score for a given dataset. The evaluation results are aggregated in different tabular formats and diagrams, formatted, and statistically analyzed (see \cref{sec:results} and \cref{sec:appendix:more_results}). While recording running time is straightforward, the quantitative evaluation of reliability and consistency requires an understanding of quality. Following previous works~\cite{van2021decaf, alaa2022how, beyond_privacy}, we break down quality into different aspects. Specifically, we consider four key aspects, which hereafter will be referred to as \emph{categories} in the evaluation:

\textit{Fidelity} refers to the similarity of individual synthetic data instances to real ones, ensuring they exhibit realistic properties such as patterns, trends, and volatility in time series. \\
\textit{Generalization} is a generator's ability to create data beyond the training data itself or noisy versions thereof. \\
\textit{Privacy} aims to reduce or even eliminate the risk of disclosure of sensitive information in the data. \\
\textit{Representativeness} is the plausibility of $D_s$ to be a sample of $P$, which can be thought of as the closeness of the synthetic dataset to the real one in the feature space. This implies an amount of diversity proportional to the real data and also extends to the utility of $D_s$ for downstream tasks.

\subsection{Determining Reliability and Consistency} To quantify reliability, we compute an indicator $r_\text{rel} \in [0,1]$ for each category and measure $m$. We define the expected category-dependent behavior of $m$ on transformation $T$ with four options: \textit{Improve} means we expect the score to get better with increasing $\kappa$; \textit{Worsen} represents the expectation of a worsening score; \textit{Constant} expects the score to remain largely unaffected by a changing $\kappa$; and \textit{N/A} means that $T$ is not applicable in this category. These expected behaviors can be confidently assigned a priori. The result is listed in \cref{tab:expected_behavior}. If the behavior is defined, we can compute a task-specific indicator $r_\text{rel}(t)$ using $m$'s scores $s_0, \dots, s_{k-1}$, where $k$ is the number of modulation steps. \\
\textit{Improve.} Assuming $s_i \in \mathbb{R}$ and improvement means to increase $s_i$, we define $r_\text{rel}$ as the fraction of score pairs $(s_i, s_j)$, $i < j$ with $s_i < s_j$. If instead $s_i$ is a boolean, the performance is determined based on a point system, where points are assigned based on the positions of \texttt{true} and \texttt{false} along the modulation path and then normalized to $[0,1]$. \\
\textit{Worsen.} If $s_i \in \mathbb{R}$, we symmetrically use the fraction of pairs with $s_i > s_j$, whereas if $s_i$ is a boolean, we swap the points assigned for \texttt{true} and \texttt{false}. \\
\textit{Constant.} For real-valued $s_i$, we expect all scores to be close to their median. We use the median instead of the mean for robustness. For boolean scores, $r_\text{rel}(t)$ is the normalized number of unequal consecutive values.
$m$'s reliability $r_\text{rel}$ is the average value across all $t$. More details and an example are provided in \cref{sec:appendix:evaluation:performance}.

 We define consistency $r_\text{con}$ of $m$ via the pairwise statistical difference between groups of $r_\text{rel}$ indicators, where each group has either the same random seed or underlying dataset. The idea is that the behavior of $m$ should depend on the relationship of synthetic and real data and not be impacted significantly by randomness or the real dataset alone. We use $r_\text{rel}$ as a proxy. Please find additional information in \cref{sec:appendix:evaluation:consistency}.

 The running times of measures and embedders are recorded separately. For each test, STEB tracks the time required for the measure to compute the score on a fully prepared dataset, that is, after transformation and embedding. If an embedding is required, the amounts of time needed for training (if applicable) and inference are tracked.
\section{Experimental Design and Execution}
\label{sec:experiments}

In this section, we outline two different experiments conducted with STEB. For the modulation, we use eleven equally spaced steps, that is, $\kappa = 0.0, 0.1, \dots, 1.0$, in both experiments. Similarly, we use all 10 datasets in each experiment. In term of hardware, we used a server with an AMD EPYC 7763, 256GB RAM, and an NVIDIA A100 running CUDA 12.4 capped at 40GB per worker. All running times were recorded on this system. To speed up the computation, we utilize up to five workers running in parallel. Still, we impose a strict time limit of 120 minutes per test to avoid excessive running times.

\subsection{Main Experiment: Ranking Measures}
\label{subsec:experiments:main}

We analyze 41 of STEB's 44 implemented measures. Of the remaining three measures, two are qualitative and thus difficult to assess objectively, while one computes instance-level scores unsuited to be condensed to a dataset-level score. Each measure is evaluated in all four categories. The goal is a comparison regarding reliability, consistency, and running time to test which measures are best suited to evaluate a specific category and within what time frame. Importantly, the goal is to provide running time estimates rather than precise measurements. We use TS2Vec as embedding and ten random seeds.

\subsection{Side Experiment: TS Embedding Models}
\label{subsec:experiments:embedders}

Embedder-dependent measures operate on embedded vector data instead of the original time series, adding an extra source of variation to the final score. This issue was previously raised in the review of \cite{huang2018an}. To this end, we examine the role of the embedding model in the synthesis evaluation process using STEB, providing empirical results for the 24 implemented embedder-dependent measures. We pairwise compare TS2Vec, Catch22, and Concat on a test-by-test basis. More specifically, we pair the tests of different embedders on the otherwise matching parameters dataset, transformation, and random seed, to compute two metrics between the two matched score sets: the mean absolute percentage error (MAPE) and the Pearson correlation coefficient (PCC). Tests for which no matching is possible due to a failed test on one side are ignored.

\section{Results}
\label{sec:results}

Given our experiment setup, the performed experiments resulted in an extensive output. Each of the 63,601 successful tests produces at least eleven scores along with multiple running time measurements. We evaluate and present them in different ways, some of which are placed in \cref{sec:appendix:more_results} due to space constraints. Not all of the 68,666 tests are successful and count when evaluating the performance of the implemented measures. The success rate heavily depends on the resource and time demands of the measure, the size of the dataset, and the transformation applied. Most measures have a success rate of around $98\%$. The details are listed in \cref{tab:statistics}. Reasons for failure include system or graphics memory overflow, measure-specific exceptions, and exceeding running time. A detailed list can be found in \cref{tab:failures}. General guidelines on how to use the results for measure selection are given in \cref{sec:appendix:guide}, which can still be used when new measures change current rankings.

{\setlength\tabcolsep{4pt}
\begin{table}[t]
    \caption{Reliability indicator overview for experiment \textit{Main} (left) and embedding comparison for the \textit{Embedders} experiment. \textsc{Left}: For each measure, we list $r_\text{rel}$ as the mean across different tests and the standard deviation in all four categories (Mean $\pm$ StD). In the upper block (all but DOMIAS), we have highlighted the best (second best) indicator in bold (by underlining). DOMIAS is listed separately as its results are based on one test only and are thus not statistically significant (see \cref{tab:statistics}). \textsc{Right}: For each of the three embedder pairs, we report the mean absolute percentage error (MAPE) and the Pearson correlation coefficient (PCC). MAPE is capped at 10 (1000 \%). For DOMAIS, no tests could be matched. All values are real numbers.}
    %\caption{}
    \label{tab:overview}
    %\label{tab:embedding_comparison}
    \vskip 1mm
    \scalebox{.765}{
    \rowcolors{2}{gray!25}{white}
    \begin{tabular}{@{\hskip0pt}cccc|c|cccccc@{\hskip0mm}}
        \toprule
        Fidelity & General- & Privacy & Represen- & Measure & \multicolumn{2}{c}{TS2Vec-Concat} & \multicolumn{2}{c}{Catch22-Concat} & \multicolumn{2}{c}{Catch22-TS2Vec} \\
         & ization &  & tativeness &  & {\footnotesize MAPE} & {\footnotesize PCC} & {\footnotesize MAPE} & {\footnotesize PCC} & {\footnotesize MAPE} & {\footnotesize PCC} \\
        \midrule
        $.416 \pm .413$ & $\textbf{.726} \pm \textbf{.320}$ & \underline{$.676$} $\pm$ \underline{$.313$} & $.347 \pm .369$ & ACS &  &  &  &  &  & \\
        $\textbf{.783} \pm \textbf{.305}$ & $.358 \pm .381$ & $.261 \pm .311$ & $\textbf{.745} \pm \textbf{.314}$ & $\alpha$-Precision & $9.005$ & $0.672$ & $1.872$ & $0.664$ & $2.838$ & $0.370$ \\
        $.460 \pm .385$ & $.244 \pm .247$ & $.324 \pm .244$ & $.539 \pm .347$ & ApEn &  &  &  &  &  &  \\
        $.052 \pm .124$ & $.530 \pm .442$ & $.666 \pm .391$ & $.098 \pm .189$ & Authenticity & $>10$ & $0.939$ & $>10$ & $0.896$ & $>10$ & $0.815$ \\
        $.083 \pm .165$ & $.624 \pm .417$ & $\textbf{.773} \pm \textbf{.306}$ & $.141 \pm .225$ & Autocorrelation &  &  &  &  &  &  \\
        $.608 \pm .435$ & $.253 \pm .397$ & $.210 \pm .349$ & $.599 \pm .425$ & $\beta$-Recall & $>10$ & $0.944$ & $>10$ & $0.891$ & $>10$ & $0.807$ \\
        $.660 \pm .193$ & $.566 \pm .203$ & $.475 \pm .102$ & $.603 \pm .170$ & C2ST &  &  &  &  &  &  \\
        $.404 \pm .271$ & $.387 \pm .258$ & $.320 \pm .197$ & $.396 \pm .259$ & CAS &  &  &  &  &  &  \\
        $.595 \pm .455$ & $.189 \pm .348$ & $.246 \pm .366$ & $.678 \pm .406$ & Context-FID & $>10$ & $0.003$ & $>10$ & $0.001$ & $>10$ & $0.915$ \\
        $.769 \pm .324$ & $.263 \pm .390$ & $.165 \pm .311$ & \underline{$.716$} $\pm$ \underline{$.360$} & Coverage & $>10$ & $0.781$ & $>10$ & $0.782$ & $>10$ & $0.623$ \\
        $.158 \pm .314$ & $.432 \pm .459$ & $.447 \pm .439$ & $.115 \pm .216$ & $\text{C}_\text{T}$ & $1.826$ & $0.807$ & $1.682$ & $0.555$ & $2.682$ & $0.751$ \\
        $.731 \pm .368$ & $.324 \pm .415$ & $.257 \pm .353$ & $.696 \pm .362$ & Density & $>10$ & $0.168$ & $>10$ & $-0.048$ & $>10$ & $-0.089$ \\
        $.641 \pm .278$ & $.504 \pm .269$ & $.438 \pm .200$ & $.590 \pm .257$ & Detection\_GMM & $>10$ & $0.487$ & $>10$ & $0.338$ & $>10$ & $0.180$ \\
        $.659 \pm .366$ & $.244 \pm .361$ & $.209 \pm .328$ & $.673 \pm .345$ & Detection\_linear & $0.101$ & $0.884$ & $0.416$ & $0.521$ & $0.431$ & $0.499$ \\
        $.739 \pm .246$ & $.418 \pm .300$ & $.333 \pm .212$ & $.703 \pm .236$ & Detection\_MLP & $>10$ & $0.290$ & $0.337$ & $0.517$ & $0.471$ & $0.243$ \\
        $.530 \pm .424$ & $.204 \pm .326$ & $.220 \pm .331$ & $.600 \pm .401$ & Detection\_XGB & $0.100$ & $0.968$ & $0.093$ & $0.981$ & $0.113$ & $0.975$ \\
        $.326 \pm .274$ & $.286 \pm .229$ & $.348 \pm .204$ & $.379 \pm .252$ & Discr. score  \\
        $.594 \pm .438$ & $.188 \pm .325$ & $.222 \pm .334$ & $.656 \pm .405$ & Distr. metric  \\
        $.576 \pm .432$ & $.215 \pm .341$ & $.266 \pm .355$ & $.650 \pm .392$ & FBCA & $>10$ & $0.577$ & $>10$ & $0.324$ & $>10$ & $0.587$ \\
        $.433 \pm .350$ & $.615 \pm .314$ & $.571 \pm .303$ & $.395 \pm .316$ & ICD  \\
        $.556 \pm .423$ & $.286 \pm .409$ & $.222 \pm .354$ & $.504 \pm .417$ & Impr. precision & $>10$ & $0.530$ & $>10$ & $0.441$ & $>10$ & $0.563$ \\
        $.715 \pm .339$ & $.290 \pm .386$ & $.190 \pm .308$ & $.691 \pm .344$ & Improved recall & $>10$ & $0.736$ & $>10$ & $0.816$ & $>10$ & $0.584$ \\
        $.697 \pm .301$ & $.336 \pm .296$ & $.310 \pm .273$ & $.683 \pm .298$ & INND  \\
        $.616 \pm .421$ & $.233 \pm .346$ & $.240 \pm .328$ & $.649 \pm .393$ & JSD & $>10$ & $0.609$ & $>10$ & $0.771$ & $>10$ & $0.378$ \\
        $.602 \pm .408$ & $.251 \pm .337$ & $.256 \pm .319$ & $.638 \pm .380$ & KLD & $>10$ & $0.396$ & $>10$ & $0.570$ & $>10$ & $0.281$ \\
        $.382 \pm .458$ & \underline{$.684$} $\pm$ \underline{$.418$} & $.524 \pm .436$ & $.231 \pm .389$ & Max-RTS & $>10$ & $0.398$ & $>10$ & $0.706$ & $>10$ & $0.461$ \\
        $.600 \pm .318$ & $.394 \pm .320$ & $.356 \pm .299$ & $.600 \pm .311$ & MTop-Div & $>10$ & $-0.006$ & $1.012$ & $0.529$ & $>10$ & $-0.020$ \\
        $.472 \pm .384$ & $.291 \pm .344$ & $.197 \pm .255$ & $.414 \pm .369$ & NDB & $>10$ & $0.559$ & $>10$ & $0.614$ & $>10$ & $0.444$ \\
        $.295 \pm .279$ & $.098 \pm .160$ & $.075 \pm .124$ & $.313 \pm .277$ & NDB-over/under & $>10$ & $0.715$ & $>10$ & $0.739$ & $>10$ & $0.597$ \\
        $.713 \pm .332$ & $.333 \pm .320$ & $.283 \pm .259$ & $.689 \pm .322$ & ONND  \\
        $.542 \pm .258$ & $.384 \pm .207$ & $.409 \pm .186$ & $.570 \pm .229$ & Predictive score  \\
        $.601 \pm .388$ & $.300 \pm .334$ & $.305 \pm .307$ & $.656 \pm .348$ & RTS & $8.306$ & $0.373$ & $>10$ & $0.839$ & $>10$ & $0.412$ \\
        $.532 \pm .440$ & $.158 \pm .265$ & $.163 \pm .263$ & $.584 \pm .428$ & Sig-MMD  \\
        $.307 \pm .348$ & $.216 \pm .252$ & $.258 \pm .256$ & $.352 \pm .346$ & Spatial corr.  \\
        $.630 \pm .415$ & $.616 \pm .383$ & $.506 \pm .362$ & $.578 \pm .398$ & STS & $>10$ & $0.256$ & $1.837$ & $0.547$ & $1.979$ & $0.209$ \\
        $.462 \pm .382$ & $.243 \pm .241$ & $.314 \pm .229$ & $.542 \pm .342$ & Temporal corr.  \\
        $.551 \pm .416$ & $.324 \pm .358$ & $.330 \pm .339$ & $.586 \pm .397$ & TRTS  \\
        $.429 \pm .309$ & $.332 \pm .228$ & $.415 \pm .184$ & $.501 \pm .271$ & TSTR  \\
        \underline{$.773$} $\pm$ \underline{$.274$} & $.448 \pm .330$ & $.337 \pm .225$ & $.713 \pm .278$ & WCS  \\
        $.617 \pm .404$ & $.237 \pm .330$ & $.233 \pm .302$ & $.647 \pm .376$ & WD & $>10$ & $0.503$ & $>10$ & $0.874$ & $>10$ & $0.457$ \\
        \midrule
        \rowcolor{white}
        $1.  \pm .000$ & $.000 \pm .000$ & $.000 \pm .000$ & $1.  \pm .000$ & DOMIAS & \multicolumn{6}{c}{--} \\
        \bottomrule
    \end{tabular}
    }
    \vspace{-4.5mm}
\end{table}
}

\subsection{Main Experiment}
\label{subsec:results:main}

The main experiment included 41,432 tests, of which 39,063 were successful. Unfortunately, for the measure \emph{DOMIAS}, all but one test failed, mostly due to excessive GPU memory demands. Still, this observed high resource demand is already a valuable insight for potential users. For the remaining 40 measures, we calculated $r_\text{rel}$, $r_\text{con}$, and the average running time. We list the reliability indicators in \cref{tab:overview} including the standard deviation observed between tests in alphabetical order. A ranking by category is shown in \cref{tab:ranking}.

The rankings for fidelity and representativeness are overall similar with \emph{$\alpha$-Precision} as the top position in both categories. \emph{ACS} and \emph{autocorrelation} perform surprisingly well in the generalization and privacy categories and not as well regarding fidelity and representativeness, which they are actually intended for. Generalization measures such as $C_T$ are further down in the ranking. More intuitive is the third position of the ``novelty'' measure \emph{authenticity} in the privacy category. Across categories, the best measures exhibit a reliability indicator of approximately $r_\text{rel} = 0.75$, which leaves room for improvement for future measures. However, $r_\text{rel}$ has a high standard deviation, meaning that closely positioned measures cannot be ordered definitively. 
% To indicate which differences are significant, we attached a critical difference diagram \cite{benavoli2016should} for each category in Figures~\ref{fig:ccd_fidelity},~\ref{fig:ccd_generalization},~\ref{fig:ccd_privacy}, and~\ref{fig:ccd_representativeness}. 
As expected, measures performing well in the categories fidelity and representativeness perform poorly in the categories generalization and privacy. This suggests that at least two measures should be used but not necessarily one for each category. Comparing the embedder-dependent and direct measures, there is no clear indication to suggest that directly applied measures are better or worse than embedder-dependent ones.

Considering the consistency results in \cref{tab:consistency}, we observe pronounced differences in the impact of the dataset and the random seed. While the measures are overwhelmingly indifferent to randomness, the reliability of a measure depends more (and sometimes heavily) on the dataset. Particularly poor in this regard are \emph{ACS}, \emph{ICD}, and \emph{density}. \emph{JSD}, \emph{KLD}, and \emph{FBCA} stand out as rather consistent measures. Most measures demonstrate mediocre consistency, varying between categories, but without clearly favoring one specifically.

We report the average running times (per dataset) for measures in \cref{tab:rt_measures} and for embedders in \cref{tab:rt_embedders}. Note that the running times for measures do not include any preparation steps such as scaling or embedding. For embedders, these comprise training (where applicable), one inference for the real data, and one for the synthetic data. The fastest to compute measures are \emph{temporal} and \emph{spatial correlation} ($\approx 0s$), which benefit from their radical subsampling approach. The measures on subsequent positions benefit from the separately recorded embedding. With the exception of \emph{autocorrelation} computed for PTB diagnostic ECG ($\approx 660s$), the execution times are negligible up to rank 27, staying below one minute. Especially the more complex, often deep-learning based measures run significantly longer, up to $8$ minutes. Naturally, Concat is the fastest embedding, followed by Catch22 typically taking under 2 minutes and TS2Vec taking up to an hour. Still, the tables only reflect part of the picture. Tests are often stopped due to excessive running times, which underestimates the actual value for some measures. 

\subsection{Embedders Experiment}
\label{subsec:results:embedders}

The results of this experiment are shown in \cref{tab:overview} on the right. Measured in MAPE, the effects of changing the embedding are remarkable. The smallest MAPE is $0.093$, while in 59 out of 61 cases it is over $100\%$, and in 51 cases it is even over $1000\%$. PCC is mostly positive and in 42 cases above $0.5$. Hence, the scores are often correlated but very different in value. Surprisingly, there is no notable difference between the TS2Vec-to-Catch22 pair (middle columns) and the comparisons with the naive Concat embedding (left and right columns). This may be due to the already big differences between TS2Vec and Catch22. Generally, there appears to be no rule or schema, neither between measures nor between embedder pairs. However, we see that the chosen embedding has an enormous effect on the score of a measure in this experiment.
\subsection{Discussion of Key Findings} 
\label{subsec:results:discussion}

In our reliability ranking, the measures \emph{$\alpha$-precision}, \emph{ACS}, \emph{autocorrelation}, and again \emph{$\alpha$-precision} take a close first place in the categories fidelity, generalization, privacy, and representativeness, respectively. However, relatively high standard deviations make a precision ranking impossible. Some positions like \emph{$\alpha$-precision}'s position in fidelity are very intuitive, others like \emph{autocorrelation}'s first place in privacy are very surprising, suggesting previously unknown properties of these measures. The side experiment on embedding models shows the influence of the chosen embedder on the measure's score, implying that generators should always be evaluated using the same embedder and motivating dedicated analysis. This motivates further research towards a suitable standardized model.

\subsection{Limitations} 
\label{subsec:results:limitations}

STEB is mainly limited by the choice and combination of transformations. While it incorporates many diverse transformation designs, their connection is not fully explored and other designs may be better suited to test certain categories. Naturally, this also limits the generalizability of the results with respect to all other potential generation methods. Furthermore, STEB currently tests four categories, while a broader range is necessary for complete coverage of desirable synthetic data properties. For instance, fairness \cite{fairness-evaluation} or splitting representativeness into diversity and utility could be investigated in future work. To accommodate new measures in the future, we designed STEB with extensibility in mind. As for the experimental design, not all potentially relevant options can be explored due to the combinatorial explosion of the number of test configurations. However, we are confident to cover a wide and deep array of parameters. 

\section{Conclusion}
\label{sec:conclusion}

Currently, the comprehensive comparison of synthetic TS quality remains challenging due to the hodgepodge of measures, little analysis of their efficacy, and generally lacking standardization in evaluation. To tackle these obstacles, we propose a novel benchmark for evaluating the performance of quality measures for synthetic time series. Our benchmark STEB computes indicators for the reliability and consistency of the scores and tracks the running time for computing each measure. To this end, we employ an array of TS transformations along a modulation path to control data modification. We utilized STEB to compare and rank 41 quantitative measures and found that the choice of TS embedding has significant impact on the measure's score. As STEB will be open-sourced after acceptance, we plan to improve and extend this benchmark with the community. This includes the handling of variable length TS, the addition of other measures, and new transformations. Lastly, it would be interesting to investigate the measures' sensitivity to the sizes of real and synthetic datasets.

{\small
\bibliography{main}
\bibliographystyle{plain}
}

\appendix
\section{Summary of Evaluation Measures}
\label{sec:appendix:measures}

Below, we list each measure analyzed in this work in alphabetical order and briefly describe it. For many of them, a detailed definition can be found in \cite{stenger2024evaluation}. In any case, we reference the original work introducing each measure as well as other sources used in their (re-)implementation. An asterisk indicates that the measure requires a TS embedding. In this case, we use vectors $\vec{x}, \vec{y}$ to represent TS $X, Y$. $l$ is the TS length, $d$ the number of feature channels, and $\delta$ the embedding dimension.

\paragraph{\acs} \cite{li2022tts-gan}. Average cosine similarity (ACS) compares all pairs of real TS $X$ and synthetic TS $Y$ with respect to their cosine similarity $\frac{\vec{x} \cdot \vec{y}}{\lvert \lvert \vec{x} \rvert \rvert \lvert \lvert \vec{y} \rvert \rvert}$. The vectors $\vec{x}$ and $\vec{y}$ are calculated by aggregating seven different statistics of $X$ and $Y$, respectively. If the data is labeled, the pairs are only constructed within each class. The final score is computed by averaging the similarities of all pairs.

\paragraph{\apen} \cite{leznik2022sok}. Approximate entropy (ApEn) was proposed by \cite{richman2000physiological} to determine the regularity and complexity of univariate time series. The synthetic data measure is derived by computing the squared difference between the approximate entropy of each channel of the real and synthetic time series. \cite{leznik2022sok} subsample both datasets to speed up the computation due to the quadratic complexity. We chose a sample size of $n=100$.

\paragraph{\authenticity*} \cite{alaa2022how}. Authenticity seeks to measure the proportion of synthetic instances that are novel relative to the real dataset by comparing the distance of the nearest synthetic neighbor to that of the nearest real one. The distances are computed in a spherical embedding space where samples closer to the origin are meant to be typical instances of the real data distribution.

\paragraph{\autocorr} \cite{ni2020conditional}. The autocorrelation measure is the squared difference of auto-correlation matrices computed for the real and synthetic dataset, respectively. The auto-correlations are determined for each channel up to lag $\frac{l}{4}$ and averaged across the TS. 

\paragraph{\ctwost} \cite{two-sample}. The classifier 2-sample test (C2ST) generally assesses whether two sets of data points are sampled from the same distribution. Here, this is realized through a DL-based binary classifier $c : X \rightarrow \{0, 1\}$ applied to real data as class $0$ and synthetic data as class $1$, combined with a hypothesis test on the class predictions. $c$ is trained on $D_\text{train}$ and the predictions taken from $D_\text{held\_out}$ and a portion of $D_s$.

\paragraph{$\textbf{C}_\textbf{T}$*} \cite{data-copying}. The data copying test targets generator overfitting, that is, it detects synthetic data that are merely minimal variations of real data instances. Using $D_\text{train}$ and $D_\text{held\_out}$, the measure compares the distances between TS from $D_\text{train}$ and $D_\text{held\_out}$ with those between TS from $D_\text{train}$ and $D_s$ using an hypothesis test. Ideally, these distances should be approximately even in both cases.

\paragraph{\cas} \cite{cas}. The classifier accuracy score (CAS) is a measure for conditional generative models, meaning it requires labeled data. This method trains a deep classifier separately on $D_\text{S}$ and $D_\text{train}$, yielding two models. Both are evaluated on $D_\text{held\_out}$ to see if the accuracies achieved by both models are similar.

\paragraph{\contextfid*} \cite{psa-gan}. Context-FID is a derivative of the ``regular'' Fr\'echet inception distance (FID) popular in image synthesis. The original implementation of Context-FID used the unsupervised representation model proposed by \cite{franceschi2019unsupervised}, while we separated the embedding step and distance calculation.  We replaced the representation model with a newer method, TS2Vec.

\paragraph{\coverage*} \cite{density-coverage}. This measure counts the number of real data instances $\vec{x}$ for which there is a synthetic instance $\vec{y}$ in its neighborhood and divides it by the size of $D_r$. Coverage $C$ is defined as
\begin{equation}
    C \coloneqq \frac{1}{\left| D_r \right|} \sum_{\vec{x} \in D_r} \textbf{1} \{ \exists \vec{y} \in D_s \colon \vec{y} \in B(\vec{x}, \text{dNN}_k(\vec{x}, D_r)) \}.
\end{equation}
where $\text{dNN}_k(\vec{x}, D)$ is the distance of vector $\vec{x}$ to the $k$th-nearest neighbor in $D$ and $B(c, r)$ a ball in the vector space with center $c$ and radius $r$.

\paragraph{\density*} \cite{density-coverage}. The density measure determines for each synthetic instance in how many neighborhoods of real instances it is located, adds the result up across all synthetic instance, and divides the sum by the neighborhood size times the size of the synthetic dataset. For the embedded real $\vec{x}$ and synthetic $\vec{y}$, density $D$ is given by
\begin{equation}
    D \coloneqq \frac{1}{K \left| D_s \right|} \sum_{\vec{y} \in D_s} \sum_{\vec{x} \in D_r} \textbf{1} \{ \vec{y} \in B(\vec{x}, \text{dNN}_k(\vec{x}, D_r)) \},
\end{equation}
where $\text{dNN}_k(\vec{x}, D)$ is the distance of vector $\vec{x}$ to the $k$th-nearest neighbor in $D$ and $B(c, r)$ a ball in the vector space with center $c$ and radius $r$.

\paragraph{\detectionmlp*} \cite{synthcity}. This is a variant of Discriminative score applied to already embedded data and with a multilayer perceptron (MLP) with depth two and 100 hidden units as discriminator model. Its score is the AUCROC score of classifying real and synthetic data.

\paragraph{\detectionxgb*} \cite{synthcity}. This is a variant of Discriminative score applied to already embedded data and with XGBoost classifier as discriminator model. Its score is the AUCROC score of classifying real and synthetic data.

\paragraph{\detectiongmm*} \cite{synthcity}. This is a variant of discriminative score applied to already embedded data and with a Gaussian mixture model (GMM) as discriminator. Its score is the AUCROC score of classifying real and synthetic data.

\paragraph{\detectionlinear*} \cite{synthcity}. This is a variant of discriminative score applied to already embedded data and with a logistic regression classifier as discriminator model. Its score is the AUCROC score of classifying real and synthetic data.

\paragraph{\discriminative} \cite{time-gan}. Uses a post-hoc RNN to classify (i.e., discriminate) original data and synthetic data with the achieved accuracy as score. We re-inplemented the original architecture, but updated the training procedure and standardized it for all DL-based discriminators.

\paragraph{\distributionalmetric} \cite{deep-hedging}. Distributional metric compares the real to the synthetic data distribution based on their probability mass function. To this end, a binning of the values in each channel across the real dataset on one, and the synthetic dataset on the other side, is performed. Finally, the mean absolute difference between binnings of corresponding channels of real and synthetic datasets is calculated and averaged over all channels.

\paragraph{\domias*} \cite{domias}. DOMIAS is a data privacy measure centered around a density-based membership inference attack. The attack aims to infer membership by targeting local overfitting of the generative model. This measure was proposed in three variants in the original work. We utilize the best-performing one which uses a block neural auto-regressive flow (BNAF) for density estimation.

\paragraph{\fbca*} \cite{coscigan}. Feature-based correlation analysis (FBCA) calculates the discrepancy of correlations of feature vectors extracted from the real and synthetic time series. It applies five different statistics to the two correlation matrices. These are mean absolute error, mean squared error, Frobenius norm, Kendall rank correlation coefficient, and Spearman rank correlation coefficient.

\paragraph{\icd} \cite{leznik2022sok} Intra-class distance (ICD) measures the average distance between the generated TS using the dynamic time-warping distance (DTW) \cite{dtw}. Formally, ICD is defined as
\begin{equation}
    \text{ICD} \coloneqq \frac{\sum_{Y \in D_s} \sum_{Y' \in D_s} \text{DTW}(Y,Y')}{\left| D_s \right|^2}.
\end{equation}
\cite{leznik2022sok} subsample $D_s$ to speed up the computation due to the quadratic complexity. We chose a sample size of $n=100$, considerably more representative than the original $n=10$.

\paragraph{\improvedprecision*} \cite{improved-p-r}. This measure counts and averages the number of generated TS for which there is a real TS in its vicinity. More formally,
\begin{equation}
    \text{IP} = \frac{1}{\left| D_s \right|} \sum_{\vec{y} \in D_s} \textbf{1} \{ \exists \vec{x} \in D_r \colon \vec{y} \in B(\vec{x}, \text{dNN}_k(\vec{x}, D_r)) \},
\end{equation}
where $\text{dNN}_k(\vec{x}, D)$ is the distance of vector $\vec{x}$ to the $k$th-nearest neighbor in $D$ and $B(c, r)$ a ball in the vector space with center $c$ and radius $r$.

\paragraph{\improvedrecall*} \cite{improved-p-r}. 
\begin{equation}
    \text{IP} = \frac{1}{\left| D_r \right|} \sum_{\vec{x} \in D_r} \textbf{1} \{ \exists \vec{y} \in D_s \colon \vec{x} \in B(\vec{y}, \text{dNN}_k(\vec{y}, D_s)) \}.
\end{equation}

\paragraph{\innd} \cite{visual}. The incomming nearest neighbor distance (INND) calculates the average dynamic time-warping distance (DTW) of any synthetic TS to its nearest real neighbor. Formally, 
\begin{equation}
    \text{INND} = \frac{1}{\lvert D_s \rvert} \sum_{Y \in D_s} \min_{X \in D_r} \text{DTW}(Y, X).
\end{equation}
\cite{visual} subsample $D_R$ and use only one synthetic TS to monitor training. To evaluate, we chose a sample size of $n=100$ for both datasets.

\paragraph{\jsd*} \cite{ouyang2018trajectory}. Generally speaking, the Jensen-Shannon divergence (JSD) measures the dissimilarity between two distributions. In this case, these are the distributions of scalar values in the TS in the real and synthetic datasets, discretized by binning these values.

\paragraph{\kld*} \cite{richardson2018on}. Similar to JSD, the Kullback-Leibler divergence measures the dissimilarity between two distributions. Again, these are the distributions of scalar values in the TS in the real and synthetic datasets, discretized by binning these values.

\paragraph{\maxrts*} \cite{norgaard2018synthetic}. The Maximum real to synthetic similarity (Max-RTS) computes the cosine similarity between the real and synthetic TS closest to each other in the embedding space given by
\begin{equation}
    \text{Max-RTS} = \max_{\vec{x} \in D_r, \vec{y} \in D_s} \left\{ \frac{\vec{x} \cdot \vec{y}}{\lvert\lvert\vec{x}\rvert\rvert_2 \cdot ||\vec{y}||_2} \right\} .
\end{equation}

\paragraph{\mtopdiv*} \cite{mtop-divergence}. Manifold topology divergence (MTop-Div) determines the discrepancy between the real and synthetic data distribution topologically. The datasets are interpreted as point clouds and topological concepts are used to assess the similarity of the two clouds.

\paragraph{\ndb*} \cite{richardson2018on}. Number os statistically different bins (NDB) assesses the degree to which the modes in the synthetic match those in the real training dataset. The modes are estimated using a K-means clustering, and matches are determined using a two-sample hypothesis test comparing pairs of real and synthetic clusters. As a baseline, the procedure is repeated with real held-out data. The score is the absolute difference between both sums of matching clusters (real-real vs real-synthetic).

\paragraph{\ndbou*} \cite{data-copying}. This is an adaption of the NDB measure. The main difference is that here, the hypothesis tested is the equality of cluster distributions. That is, the distributions of corresponding real and synthetic clusters should be the same. The goal is to detect under-represented and over-represented data regions. The final score is two-fold: A real number for the number of under-represented clusters, and one for the number of over-represented clusters.

\paragraph{\onnd} \cite{visual}. The outgoing nearest neighbor distance (ONND) calculates the average dynamic time-warping distance (DTW) of any real TS to its nearest synthetic neighbor. Formally, 
\begin{equation}
    \text{ONND} = \frac{1}{\lvert D_r \rvert} \sum_{X \in D_r} \min_{Y \in D_s} \text{DTW}(X, Y).
\end{equation}
\cite{visual} subsample $D_R$ and $D_s$ to speed up the computation due to the quadratic complexity. We chose a sample size of $n=100$.

\paragraph{\predictive} \cite{time-gan}. This measure trains a simple forecasting model on the generated time series to conduct one-step-ahead predictions. Then, it evaluates its performance on real data using mean absolute error (MAE) and returns its mean.

\paragraph{\rts*} \cite{norgaard2018synthetic} Real-to-synthetic similarity (RTS) compares the average cosine similarity within the real data to the cosine similarity between all real TS and $10$ random synthetic TS. More, specifically, the score is given by
\begin{equation}
    \text{RTS} = \left| \frac{1}{10 \cdot \lvert D_r \rvert} \sum_{i=1}^{10} \sum_{\vec{x} \in D_r} \frac{\vec{x} \cdot \vec{y}_i}{||\vec{x}||_2 \cdot ||\vec{y}_i||_2} - \binom{|D_r|}{2} \sum_{i \not = j} \frac{\vec{x}_i \cdot \vec{x}_j}{|| \vec{x}_i ||_2 \cdot ||\vec{x}_j||_2} \right|.
\end{equation}

\paragraph{\sigmmd} \cite{lou2023pcf}. Computes the maximum mean discrepancy (MMD) between signature features extracted from the real and synthetic dataset, respectively. We use a signature kernel with random fourier features (RFF) map and tensor random projections (TRP) from the KSig library\footnote{\url{https://github.com/tgcsaba/KSig}}.

\paragraph{\spatial} \cite{leznik2022sok}. Spatial correlation calculates the squared difference of inter-channel Pearson correlations of multivariate real and synthetic TS. For each time series and on both datasets separately, the correlation coefficient is determined for each pair of channels and averaged within the dataset. To reduce computational cost, both datasets are sampled down to $n=100$ TS.

\paragraph{\sts*} \cite{norgaard2018synthetic}. Synthetic-to-synthetic similarity measures the ``typical'' cosine similarity between embedded synthetic TS. Typical means that for every instance, the distances to only five others are calculated. This measure does not use real data.

\paragraph{\temporal} \cite{leznik2022sok}.  Computes the channel-wise correlation between observations of each TS using the frequency peaks exposed by the fast Fourier transformation (FFT). Then, calculates the mean squared difference between these correlations in the real and synthetic dataset.

\paragraph{\trts} \cite{rcgan}. ``Train on real-test on synthetic'' measures how well a model trained on real data performs on the generated data. The absolute difference between the performances on real data versus on synthetic data is used as score. For forecasting, the same model as in predictive score \cite{time-gan} is used.

\paragraph{\tstr} \cite{rcgan}. ``Train on synthetic-test on real'' is a utility-based measure determining the usefulness of the synthetic data on a downstream task compared to real data. As downstream task, we also choose forecasting and utilize the model used for predictive score \cite{time-gan}. A score is computed by taking the absolute performance difference between the forecaster trained on synthetic data and the one trained on real data, evaluated on held-out real test data. This last piece sets the two measures apart.

\paragraph{\wcs} \cite{tts-cgan}. The wavelet coherence score (WCS) computes the mean pairwise wavelet coherence, which is an analysis technique for time and frequency correlation, between real and synthetic TS. To reduce computational cost, both datasets are sampled down to $n=100$ TS. 

\paragraph{\wsd*} \cite{richardson2018on}. The Wasserstein-1 distance (WD) is a distance function between probability distributions. In this case, it is applied to the scalar values of the TS in the real and synthetic dataset. For each set a discrete $1$D distribution is created by binning the values. The WD is applied to the real and synthetic binning.

\paragraph{\alphaprecision*} \cite{alaa2022how}. This measure is based around a fraction $\alpha \in [0, 1]$ of real time series considered ``typical'' for this data distribution. A synthetic TS falling into the support of the typical part of the real distribution is considered realistic and faithful. A score is deduced by aggregating the deviation between expected fraction of synthetic TS in the support and the actual fraction over different values for $\alpha$. The support is determined in a spherical embedding space where samples closer to the origin are meant to be typical instances of the real data distribution.

\paragraph{\betarecall*} \cite{alaa2022how}. Analogous to $\alpha$-Precision, this measure is based around a fraction $\beta \in [0, 1]$ of synthetic time series considered ``typical'' for this generator. For each $\beta$, the measure determines the fraction of real TS with at least one typical synthetic TS in its vicinity. A score is derived by taking the average of the divergence between the expected fraction and the actual one across the values for $\beta$. The distances are computed in a spherical embedding space where samples closer to the origin are meant to be typical instances of the real data distribution.

The following three measures are part of the benchmark, but were not used in any of the experiments:

\paragraph{\distvisual} \cite{time-gan}. This is a qualitative measure, creating a dot-plot visualizing the distributions of the embedded real and synthetic data. The embedding is computed with t-distributed stochastic neighbor embedding (t-SNE) or principal component analysis (PCA) of a subsample of $n=1000$ instances each.

\paragraph{\visual} \cite{visual}. This measure is based on the visual evaluation of generated time series data by plotting a (small) subsample of the synthetic dataset. Ideally, the plots are assessed by multiple domain expert.

\paragraph{\realism*} \cite{improved-p-r}  The measure approximates instance fidelity via the position of an embedded synthetic TS in the real data manifold. The closer the generated TS is to a real TS compared to other real TS, the more realistic the generated TS is. The Euclidean distance is used for distance calculation. Realism is a sample-level variant of improved precision and therefore not adequate for our dataset-level experiments. Instead, we test improved precision.

\section{Details on Datasets, Embedders, and Randomness}

\subsection{Datasets}
\label{subsec:appendix:datasets}

In both experiments, we used the following ten datasets. This selection covers multiple source domains, a wide range of dataset sizes, TS lengths,  and TS dimensions, as well as labeled and unlabeled data. In addition, the values themselves are diverse with respect to three statistical characteristics. Please find a summary of key characteristics in \cref{tab:dataset_statistics}.

{\setlength\tabcolsep{4pt}
\begin{table}[t]
    \caption{Summary of dataset statistics. For each dataset, this table includes its domain and size, the length and dimension of the contained TS, the number of classes, the average singular value decomposition (SVD) entropy~\cite{vallat2021antropy}, the average permutation entropy~\cite{bandt2002permutation}, and the average correlation between TS features.}
    \label{tab:dataset_statistics}
    \centering
    \scalebox{.88}{
    \rowcolors{2}{gray!25}{white}
    \begin{tabular}{@{\hskip0pt}lcccccccc@{\hskip0pt}}
        \toprule
        Dataset & Domain & Size & TS & TS & Classes & $\varnothing$ SVD & $\varnothing$ Permuta- & $\varnothing$ Feature \\
         &  &  & length & dimension &  & entropy & tion entropy & correlation \\
        \midrule
        Appliances energy & Smart home & 19592 & 144 & 28 & - & 0.265 & 1.701 & 0.044 \\
        ElectricDevices & Devices & 16637 & 96 & 1 & 7 & 1.392 & 1.343 & - \\
        Exchange rate & Finance & 7559 & 30 & 8 & - & 0.044 & 2.293 & 0.304 \\
        Google stock & Finance & 3662 & 24 & 6 & - & 0.271 & 2.311 & 0.624 \\
        PPG and respiration & Medical & 21600 & 125 & 5 & - & 0.409 & 1.670 & -0.106 \\
        PTB diagnostic ECG & Medical & 57618 & 1000 & 15 & 11 & 0.271 & 2.302 & 0.098 \\ 
        Sine & - & 10000 & 100 & 2 & 5 & 0.348 & 1.583 & 0.524 \\
        StarLightCurves & Sensor & 9236 & 1024 & 1 & 3 & 0.093 & 1.059 & - \\
        UniMiB SHAR & Motion & 11771 & 151 & 3 & 17 & 1.068 & 2.229 & -0.004 \\ 
        Wikipedia web traffic & Networking & 117277 & 550 & 1 & - & 0.497 & 2.522 & - \\
        \bottomrule
    \end{tabular}
    }
\end{table}
}

\textbf{Appliances energy} The UCI Appliances energy prediction dataset consists of multivariate measurements recorded by sensors in a low-energy building, augmented by weather readings and two random variables~\cite{energy-dataset}. Measurements were taken at 10-minute intervals for approximately $4.5$ months. By sliding a window of 144 steps with stride one along the time axis, we create a set of overlapping, individual, multivariate time series.

\textbf{ElectricDevices} This dataset is part of the UCR Time Series Classification Archive~\cite{UCRArchive2018}, which comprises 128 labeled subsets from different domains and with different characteristics. We chose ElectricDevices as the best fit with the other sets in our selection.

\textbf{Exchange rate} A collection of the daily exchange rates for the eight currencies of Australia, British, Canada, Switzerland, China, Japan, New Zealand, and Singapore, respectively, ranging from 1990 to 2016~\cite{exchange_rate}. We apply the sliding window approach again with stride one.

\textbf{Google stock} This set contains the daily historical Google stocks data from 2004 to 2019 in one continuous, aperiodic sequence with features volume, high, low, opening, closing, and adjusted closing prices~\cite{time-gan}. We apply the sliding window approach again with stride one.

\textbf{PPG and respiration} Assembled by the Beth Israel Deaconess Medical Centre(BIDMC), this dataset contains physiological signals and static features extracted from the much larger MIMIC-II matched waveform database~\cite{bidmc, physionet}. We extracted five dynamic features plus labels from the $45$ patients for which they are provided: RESP, PLETH, V, AVR, and II. The sequence length of $125$ corresponds to 1s of recordings (Sampling rate $125$Hz).

\textbf{PTB diagnostic ECG}  The PTB diagnostic ECG database is a collection of 549 15-lead ECGs (i.e., 15 feature channels) for 294 patients, including clinical summaries for each record~\cite{ptb, physionet}. We extract subsequences of 1000 steps, which corresponds to 1s-long recordings (Sampling rate $1000$Hz). Further, we use the eleven diagnosed conditions as class labels.

\textbf{Sine} This is a self-crafted dataset of time series composed of two sine waves each. The set contains multiple, imbalanced classes which differ in wave amplitude, x-shift, phase length, and phase offset between feature channels.

\textbf{StarLightCurves} This is the second dataset from the UCR Time Series Classification Archive~\cite{UCRArchive2018}. The TS are labeled.

\textbf{UniMiB SHAR} Researchers from the University of Milano-Bicocca created this mulitvariate dataset by collecting acceleration samples acquired with an Android smartphone~\cite{unimib-shar-dataset}. The three features represent X-, Y-, and Z-coordinates. Each instance is labeled with one of 17 activities, which we use as classes.

\textbf{Wikipedia web traffic} This set contains visitation data for over $100,000$ Wikipedia articles~\cite{wwt-dataset}. Each of the TS included represents the number of daily views of a different Wikipedia article, starting from July 1st, 2015 up until December 31st, 2016. The data was originally compiled for a competition with training and test data. We, however, only use the train set.

\subsection{Embedders}
STEB currently offers two non-trivial embedding models, which we also used for our experiments. These are:

\textbf{TS2Vec} is a deep-learning model for time series embedding \cite{ts2vec}. The model features a CNN-based encoder consisting of cascading dilated convolutional blocks. Training is conducted via hierarchical contrasting loss, which is crucial to the method’s success. The learned sequence representations are aggregated from representations of individual time steps created by the convolution blocks.

\textbf{Catch22.} The second model is the feature extractor catch22 \cite{lubba2019catch22}. It computes a diverse set of 24 statistical descriptors of a given univariate time series or one feature channel of a multivariate time series. For the latter, we concatenate the feature vectors of each channel to obtain an embedding for the entire time series, i.e., 
\begin{equation}
 X \mapsto \text{catch22}(\vec{c}_0) \mathbin\Vert \text{catch22}(\vec{c}_1) \mathbin\Vert \dots \mathbin\Vert \text{catch22}(\vec{c}_{d-1}), 
\end{equation}
where $d$ is the number of channels.

\subsection{Randomness}
In our experiments, randomness plays a role at different stages, mainly in splitting the real dataset after preprocessing, while training the TS2Vec embedder, and during the execution of measures. During one test run, we use the same random seed, specified in the test parametrization, at every step of the test to ensure reproducibility. For the \textit{Main} experiment, the ten seeds tried are 42, 461900, 854324, 679123, 107460, 952343, 580127, 893234, 560239, and 501932. Due to time constraints, the \textit{Embedders} experiment is limited to a subset of five seeds, namely 952343, 580127, 893234, 560239, and 501932.

\section{Additional Benchmark Details}
\label{sec:appendix:benchmark}

Below, we describe the supporting components of STEB in more detail. Their context within STEB and the relationship between components is visualized in \cref{fig:design}.

\textbf{Configuration \& Management.} This component guides and coordinates the experiment, starting with loading and validating the configuration, creating the parameter sets of the tests to run, and initiating the test processing. It monitors the execution and triggers recovery if necessary. Experiments can be flexibly executed in two modes of operation: sequential and parallel. In sequential mode, the tests are processed one after the other; caching and logging are done in the file system. The advantage is low computational overhead and few additional dependencies. In parallel mode, the component spawns a user-specified number of worker instances that select tests from a pool and process them in parallel. Users can also limit CPU and RAM use and choose between GPU-enabled and CPU-only workers. Technically, these workers are Docker containers, which implies additional dependencies to run STEB but speeds up the processing and facilitates (dependency) isolation of the measures to be evaluated.

\textbf{Storage.} In the simple sequential mode, everything is stored in a workspace in the file system. In parallel mode, the monitoring, logging, caching, handling of results, and evaluation is optimized using a MongoDB\footnote{\url{https://www.mongodb.com}} database. This is also more user-friendly, as database monitoring tools support easy, visual tracking of the experiment progress, particularly failed tests. 

\textbf{Caching.} Many artifacts such as processed datasets, trained models, or distance matrices produced by preprocessing, embedders, and measures, are duplicated across the modulation steps inside a test and throughout the different tests of an experiment. To speed up the running time, conserve valuable resources, and save energy, caching can be enabled. When it is enabled, each artifact is created once, stored away, and loaded whenever needed---provided that the test parameters match.

\textbf{Recovery.} Since individual tests sometimes fail, workers crash, and experiments are interrupted, this component handles the return to a valid program state. This includes seamlessly continuing experiments, automatically restarting workers, and cleaning up inconsistent caches.

\textbf{Logging.} This component logs various aspects of the experiment execution, informing the user and providing transparency. It records the parameter set for each test and tracks the live status of individual tests, which can be waiting (\textit{todo}), \textit{ongoing}, \textit{successful}, or \textit{failed}. Additionally, status information is recorded, such as the reason for failure. Furthermore, the beginning and end of each test processing are logged and (optionally) the running times for each embedder and measure invocation are saved. 

\section{Details of the Evaluation Procedure}
\label{sec:appendix:evaluation}

In the two subsections below, we outline the mathematical definitions of reliability $r_\text{rel}$ and consistency $r_\text{con}$. An overview of the expected behavior of each measure used to calculate $r_\text{rel}$ is provided in \cref{tab:expected_behavior}.

\begin{table}
    \caption{Summary of the expected measure behaviors per category in each test. We differentiate four major behaviors: The score improves ($\nearrow$), worsens ($\searrow$), remains constant (c), and not applicable (-). Transformations marked with * are preceded by a shuffle of the input dataset, even for $\kappa=0$.}
    \label{tab:expected_behavior}
    \vskip 0.1in
    \centering
    \begin{small}
    \rowcolors{2}{gray!25}{white}
	\begin{tabular}{lcccccc}
		\toprule
		                          & Fidelity   &  General-  & Privacy    & Represen-  \\
		                          &            &  ization   &            & tativeness \\
        \midrule                                                                                                
		Label corruption        & $\searrow$ & c          & -          & $\searrow$ \\
		Gaussian noise*         & $\searrow$ & $\nearrow$ & $\nearrow$ & $\searrow$ \\
		Misaligning channels*   & $\searrow$ & c          & $\nearrow$ & $\searrow$ \\
		Mode dropping*          & c          & c          & $\nearrow$ & $\searrow$ \\
		Mode collapse*          & c          & c          & $\nearrow$ & $\searrow$ \\
		Moving average*         & $\searrow$ & $\nearrow$ & $\nearrow$ & $\searrow$ \\
		Rare event sensitivity* & c          & c          & $\nearrow$ & $\searrow$ \\
		Reverse substitution*   & c          & $\searrow$ & $\searrow$ & c          \\
		Salt \& Pepper*         & $\searrow$ & $\nearrow$ & $\nearrow$ & $\searrow$ \\
		Segment leaking*        & $\searrow$ & $\searrow$ & $\searrow$ & $\searrow$ \\
		STL*                    & $\searrow$ & $\nearrow$ & $\nearrow$ & $\searrow$ \\
		Substitution*           & c          & $\nearrow$ & $\nearrow$ & c          \\
		Wavelet transform*      & $\searrow$ & $\nearrow$ & $\nearrow$ & $\searrow$ \\
		\bottomrule
	\end{tabular}
    \end{small}
    \vskip -0.1in
\end{table}

\subsection{Calculating the Reliability Indicator}
\label{sec:appendix:evaluation:performance}

Below, we provide the formal definition of $r_\text{rel}(t)$ for task $t$ with scores $s_0, \dots, s_{k-1}$.

\paragraph{Improve, real.} 
\begin{equation}
    r_\text{rel}(t) = \frac{2}{k(k-1)} \sum_{i=0}^{k-2} \sum_{j=i+1}^{k-1} \textbf{1}\{ s_i < s_j \}
\end{equation}

\paragraph{Improve, boolean.} Let $w^t = (w_0^t, \dots, w_{k-1}^t)^T$, $w^f = (w_0^f, \dots, w_{k-1}^f)^T$ with $w_i^t = i, w_i^f = k-i-1$ be weight vectors to assign points to the scores based on their position on the modulation path. We define
\begin{equation}
    r_\text{rel}(t) = \frac{r_\text{nominal} - r_\text{min}}{r_\text{max} - r_\text{min}}
\end{equation}
for
\begin{equation}
    r_\text{nominal} = \sum_{i=0}^{k-1} w^t_i \cdot \textbf{1}\{s_i\} + w^f_i \cdot \textbf{1}\{\lnot s\}
\end{equation}
and
\begin{equation}
    r_\text{min} = \sum_{i=0}^{\intersect-1} w_i^t + \sum_{i=\intersect}^{k-1} w_i^f, \quad r_\text{max} = \sum_{i=0}^{\intersect-1} w_i^f + \sum_{i=\intersect}^{k-1} w_i^t.
\end{equation}

\paragraph{Worsen, real.} Analogous to \textit{improve} with $>$.

\paragraph{Worsen, boolean.} Analogous to \textit{improve} with $w_i^t = k-i-1, w_i^f = i$ and
\begin{equation}
    r_\text{min} = \sum_{i=0}^{\intersect-1} w_i^f + \sum_{i=\intersect}^{k-1} w_i^t, \quad r_\text{max} = \sum_{i=0}^{\intersect-1} w_i^t + \sum_{i=\intersect}^{k-1} w_i^f.
\end{equation}

\paragraph{Constant, real.} Let $\mu$ be the median score of $m$ on $t$ and $\epsilon = 0.05$. We define
\begin{equation}
    r_\text{rel}(t) = \frac{1}{k-1} \left | \left \{ (s_i, \mu) \mid \left|\frac{\left|\mu-s_i\right|}{\mu}\right| \le \epsilon \land s_i \not = \mu \right \} \right |
\end{equation}

\paragraph{Constant, boolean.} Similar to real-valued $s_i$, for boolean values, we have
\begin{equation}
    r_\text{rel}(t) = \frac{1}{k-1} \left | \left\{(s_i, s_{i+1}) \mid \text{XNOR}(s_i, s_{i+1}), 0 \le i < k-1 \right\} \right |.
\end{equation}

\paragraph{Example.} Assume a very small experiment of arbitrary measure $m$ with just two tests $t_0, t_1$, resulting in eleven real-valued scores each, 
\begin{equation}
    s = [0, 0, 1, 2, 4, 3, 5, 6, 7, 8, 7] \quad \text{and} \quad \sigma = [2.8, 3.0, 2.9, 2.9, 3.0, 3.0, 3.1, 3.0, 3.2, 3.1, 3.0].   
\end{equation}
$s$ was calculated for transformation misalignment, $\sigma$ for mode dropping. The value at position $0$ was calculated with $\kappa = 0$, position $1$ with $\kappa = 0.1$, etc. We will determine $r_\text{rel}$ for category fidelity. According to \cref{tab:expected_behavior}, we expect diminishing fidelity as the misalignment of channels increases, i.e., $s$ to get worse. Hence, we follow paragraph \textbf{Worsen, real} and calculate
\begin{equation}
    r_\text{rel}(t_0) = \frac{2}{11(11-1)} \sum_{i=0}^{11-2} \sum_{j=i+1}^{11-1} \textbf{1}\{ s_i > s_j \} = 0.036.
\end{equation}
On the other hand, we expect approximately constant fidelity of the remaining samples even if modes collapse. We calculate
\begin{equation}
    r_\text{rel}(t_1) = \frac{1}{11-1} \left | \left \{ (s_i, 3) \mid \left|\frac{\left|3-s_i\right|}{3}\right| \le 0.05 \land s_i \not = 3 \right \} \right | = 0.8
\end{equation}
based on \textbf{Constant, real}.
Hence,
\begin{equation}
    r_\text{rel} = \frac{r_\text{rel}(t_0) + r_\text{rel}(t_1)}{2} = 0.418,
\end{equation}
attesting $m$ a mediocre to bad reliability in fidelity.

\subsection{Calculating the Consistency Indicator}
\label{sec:appendix:evaluation:consistency}

Now, the computation of consistency $r_\text{con}$ is as follows, starting with the consistency regarding a changing random seed. First, group the $r_\text{rel}(t)$ by random seed for all $t$ in the experiment. Assuming there are $n$ random seeds, we have $G_0, \dots, G_{n-1}$. Then, we apply the Kolmogorov–Smirnov test for two samples\cite{hodges1958significance} to pairs $(G_i, G_j), i < j$, count the pairs which were identified as following the same distribution, and normalize it by the number of pairs:

\begin{equation}
    r_\text{con} = \frac{2}{n(n-1)} \sum_{i=0}^{n-2} \sum_{j=i+1}^{n-1} \textbf{1}\{ \text{ks\_2sample}(G_i, G_j) \}.
\end{equation}
The consistency w.r.t. a changing dataset is analogous, just replace ``random seed'' by ``dataset'' above.

\section{Details of the Experiments}
\label{sec:appendix:experiments}

In this section, we provide further details on the experiments conducted, \textit{Main} and \textit{Embedders}. Listing~\ref{lst:config} contains the configuration for \textit{Main}. Parameters in the lower part are used for management purposes and influence only how the tests are executed, not what they do. Statistics computed for both experiments, including the number of tests and their success rate, can be found in \cref{tab:statistics}. The reasons for failed tests, on the other hand, are listed in \cref{tab:failures}. They range from excessive running time and (GPU) memory overflow to measure-specific errors that can occur in its normal operation.

\begin{lstlisting}[style=yaml, caption={This is the config file for the \textit{Main} experiment. It specifies the setup of an experimental run, including which datasets to use (here: all ten), the transformations, embedders, and measures. Transformations can be nested once, i.e., they can be chained within one test and are sequentially applied. For more details, especially the other parameters, please see the documentation of the referenced repository.}, label=lst:config]
name: main
datasets: ALL
transformations: [
  [shuffle, gn_moderate],
  [shuffle, salt_and_peper_noise],
  [shuffle, misalignment],
  [shuffle, substitution],
  [shuffle, mode_dropping],
  [shuffle, mode_collapse],
  [shuffle, reverse_sub_clean],
  corrupt_labels,
  [shuffle, segment_leaking],
  [shuffle, rare_event_drop],
  [shuffle, moving_average],
  [shuffle, stl_decomposition],
  [shuffle, wavelet_transform],
]
embedders: [ts2vec]
measures: [
  icd, ap_en, innd, onnd, spatial, temporal, c_t, sts, max_rts, rts, 
  jsd, kld, wd_on_pmf, auto_corr, wcs, ndbou, ndb, m_top_div, 
  Coverage, Density, improved_precision, improved_recall, 
  distributional_metric, context_fid, discriminative, predictive, 
  detection_mlp, detection_xgb, detection_gmm, detection_linear, 
  tstr, trts, cas, c2st, alpha_precision, beta_recall, authenticity, 
  acs, fbca, domias, sig_mmd
]
seeds: [42, 461900, 854324, 679123, 107460, 
        952343, 580127, 893234, 560239, 501932]
data_feeds: [feed_train]
use_cache: true
workers: {cpu: [[4, 100], [4, 100]], 
          gpu: [[4, 100], [4, 100], [4, 100]]}
use_database: true
rebuild_image: false
record_runtime: true
restart_failed: false
test_time_limit: 120
compare_results_to: {embedding: []}
\end{lstlisting}

\begin{table}[t]
    \centering
    \caption{Statistics for the experiments: For each measure, this table lists the total number of tests attempted, the number of successful tests, and the success rate. The totals across all measures is provided at the table bottom. Only embedding-dependent measures were tested in the \emph{embedders} experiment, entries for other measures are marked as not applicable (N/A).}
    \label{tab:statistics}
    \vskip 0.1in
    {\small
    \rowcolors{3}{white}{gray!25}
    \begin{tabular}{@{\hskip3pt}lcccccc@{\hskip3pt}}
        \toprule
                & \multicolumn{3}{c}{Main}     & \multicolumn{3}{c}{Embedders} \\
        Measure & \#Total & \#Succ & \%Succ & \#Total & \#Succ & \%Succ \\
        \midrule
        $\alpha$-Precision & 1020 & 998 & 98 & 1190 & 1149 & 97 \\
        $\beta$-Recall & 1020 & 1000 & 98 & 1171 & 1119 & 96 \\
        $\text{C}_\text{T}$ & 1020 & 988 & 97 & 1093 & 922 & 84 \\
        ACS & 1020 & 1000 & 98 & \multicolumn{3}{c}{N/A} \\
        ApEn & 1020 & 1000 & 98 & \multicolumn{3}{c}{N/A} \\
        Authenticity & 1020 & 998 & 98 & 1197 & 1152 & 96 \\
        Autocorrelation & 1020 & 1000 & 98 & \multicolumn{3}{c}{N/A} \\
        C2ST & 1020 & 996 & 98 & \multicolumn{3}{c}{N/A} \\
        CAS & 631 & 619 & 98 & \multicolumn{3}{c}{N/A} \\
        Context-FID & 1020 & 999 & 98 & 1159 & 895 & 77 \\
        Coverage & 1020 & 1000 & 98 & 1100 & 1073 & 98 \\
        DOMIAS & 1020 & 1 & 0 & 1038 & 89 & 9 \\
        Density & 1020 & 999 & 98 & 1088 & 1064 & 98 \\
        Detection\_GMM & 1020 & 998 & 98 & 1061 & 921 & 87 \\
        Detection\_MLP & 1020 & 996 & 98 & 1190 & 1136 & 95 \\
        Detection\_XGB & 1020 & 1000 & 98 & 1083 & 1042 & 96 \\
        Detection\_linear & 1020 & 998 & 98 & 1086 & 1063 & 98 \\
        Discriminative score & 1020 & 996 & 98 & \multicolumn{3}{c}{N/A} \\
        Distr. metric & 1020 & 999 & 98 & \multicolumn{3}{c}{N/A} \\
        FBCA & 1020 & 1000 & 98 & 1440 & 1424 & 99 \\
        ICD & 1020 & 1000 & 98 & \multicolumn{3}{c}{N/A} \\
        INND & 1020 & 1000 & 98 & \multicolumn{3}{c}{N/A} \\
        Improved precision & 1020 & 999 & 98 & 1184 & 957 & 81 \\
        Improved recall & 1020 & 999 & 98 & 1186 & 967 & 82 \\
        JSD & 1020 & 999 & 98 & 1078 & 1056 & 98 \\
        KLD & 1020 & 996 & 98 & 1060 & 1038 & 98 \\
        MTop-Div & 1020 & 1000 & 98 & 1209 & 1160 & 96 \\
        Max-RTS & 1020 & 999 & 98 & 1164 & 1040 & 89 \\
        NDB & 1020 & 991 & 97 & 1071 & 1014 & 95 \\
        NDB-over/under & 1020 & 924 & 91 & 1076 & 1053 & 98 \\
        ONND & 1020 & 1000 & 98 & \multicolumn{3}{c}{N/A} \\
        Predictive score & 1020 & 999 & 98 & \multicolumn{3}{c}{N/A} \\
        RTS & 1021 & 999 & 98 & 1158 & 1112 & 96 \\
        STS & 1020 & 998 & 98 & 1074 & 1053 & 98 \\
        Sig-MMD & 1020 & 691 & 68 & \multicolumn{3}{c}{N/A} \\
        Spatial correlation & 1020 & 998 & 98 & \multicolumn{3}{c}{N/A} \\
        TRTS & 1020 & 992 & 97 & \multicolumn{3}{c}{N/A} \\
        TSTR & 1020 & 998 & 98 & \multicolumn{3}{c}{N/A} \\
        Temporal correlation & 1020 & 1000 & 98 & \multicolumn{3}{c}{N/A} \\
        WCS & 1020 & 880 & 86 & \multicolumn{3}{c}{N/A} \\
        WD & 1020 & 997 & 98 & 1078 & 1058 & 98 \\
        \midrule
        \rowcolor{white}
        Total & 41432 & 39044 & 94 & 27234 & 24557 & 90  \\
        \bottomrule
    \end{tabular}
    }
\end{table}

\begin{table}[t]
    \caption{Reasons for test failure and the number of such failures for each experiment. Note that the overall number of tests also varies from experiment to experiment.}
    \label{tab:failures}
    \centering
    \scalebox{.89}{
    \rowcolors{2}{gray!25}{white}
    \begin{tabular}{lcc}
        \toprule
        Failure & Main & Embedders \\
        \midrule
        CUDA Out of Memory & 1061 & 1369 \\
        Memory limit of 100 GB exceeded & 7 & 123 \\
        Time limit of 120 minutes exceeded & 1037 & 707 \\
        NDB-over/under: Too many cells in partition/No samples in a cell & 81 & 0 \\
        Other CUDA/CUDNN Runtime error & 0 & 0 \\
        Non-CUDA Runtime error & 0 & 0 \\
        DOMIAS: BNAF density estimator produced illegal values. & 191 & 114 \\
        $C_T$: Cell $x$ is missing test or training samples. & 9 & 105 \\
        Context-FID: Imaginary component in fr\'echet distance calculation & 0 & 249 \\
        Detection\_GMM: Fitting the mixture model failed & 0 & 10 \\
        Spatial correlation: Cannot compute Pearson correlation for any of the given samples & 2 & 0 \\
        \bottomrule
    \end{tabular}
    }
\end{table}

\section{Complementary Results}
\label{sec:appendix:more_results}

Additional results for the experiments can be found here across different tables and figures. In \cref{tab:ranking}, we provide a reliability ranking of the measures in the four categories. This is an alternative presentation of the information in \cref{tab:overview}, which is alphabetically sorted. Here, we see more clearly how close the reliability indicators of neighboring measures are. For lack of space in the main body of the paper, we report the consistency indicators in \cref{tab:consistency}. They are discussed in \cref{subsec:results:main}. Similarly, the average running times recorded for measure executions are placed in \cref{tab:rt_measures}, those for embedding procedures in \cref{tab:rt_embedders}. There are also long versions for both tables, \cref{tab:rt_measures_long1}, \cref{tab:rt_measures_long2}, and \cref{tab:rt_embedders_long}, which additionally include the standard deviation, number of valid measurements, and number of invalid measurements for each measure/embedder-dataset-combination.

\begin{landscape}
\begin{figure}
    \centering
    \includegraphics[width=\linewidth, trim={0 0 0 8mm}, clip]{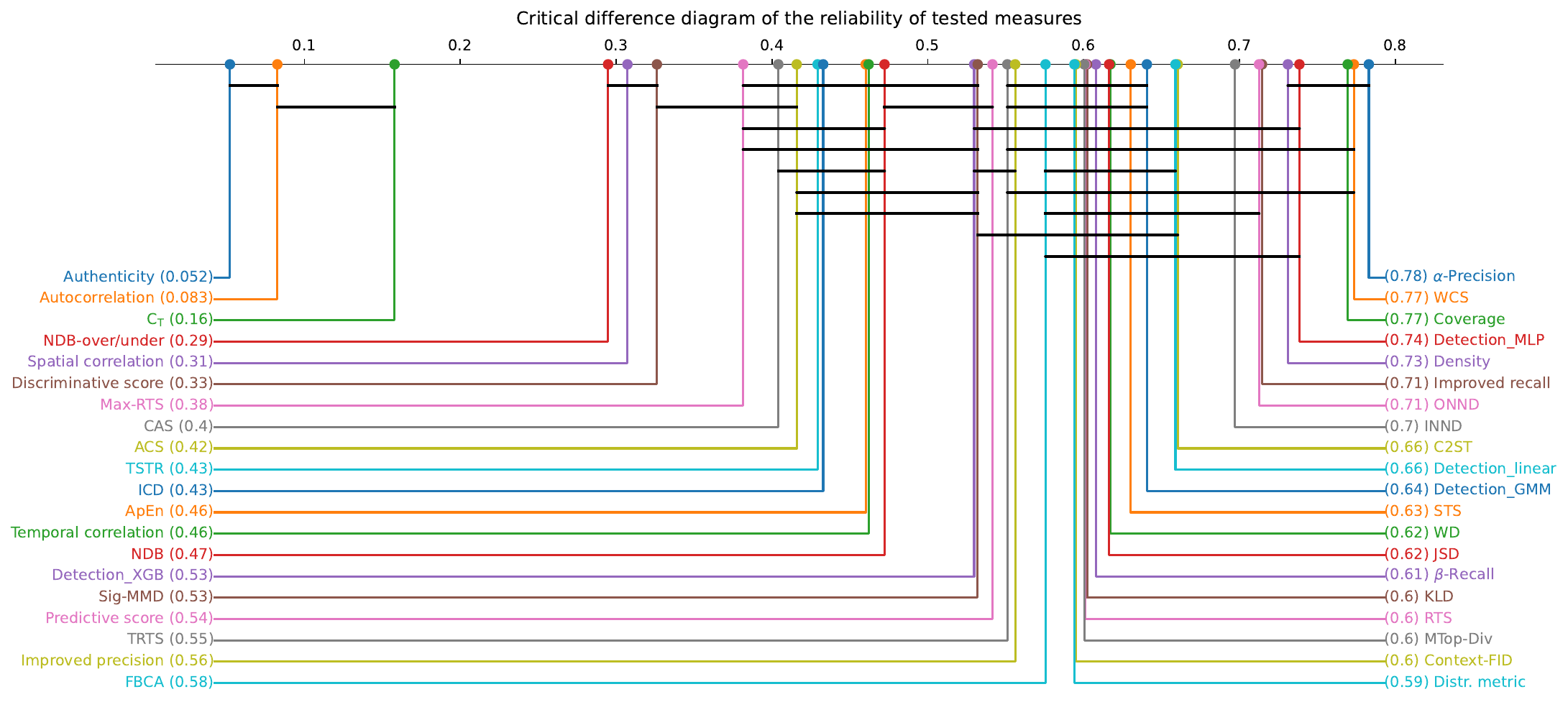}
    \caption{Critical difference diagram for reliability indicator $r_\text{rel}$ in category fidelity as part of \textit{Main}. The horizontal axis at the top depicts $r_\text{rel}$. Additional horizontal bars connect groups of measures with no significantly different $r_\text{rel}$ value.}
    \label{fig:cdd_fidelity}
\end{figure}
\end{landscape}

\begin{landscape}
\begin{figure}
    \centering
    \includegraphics[width=\linewidth, trim={0 0 0 8mm}, clip]{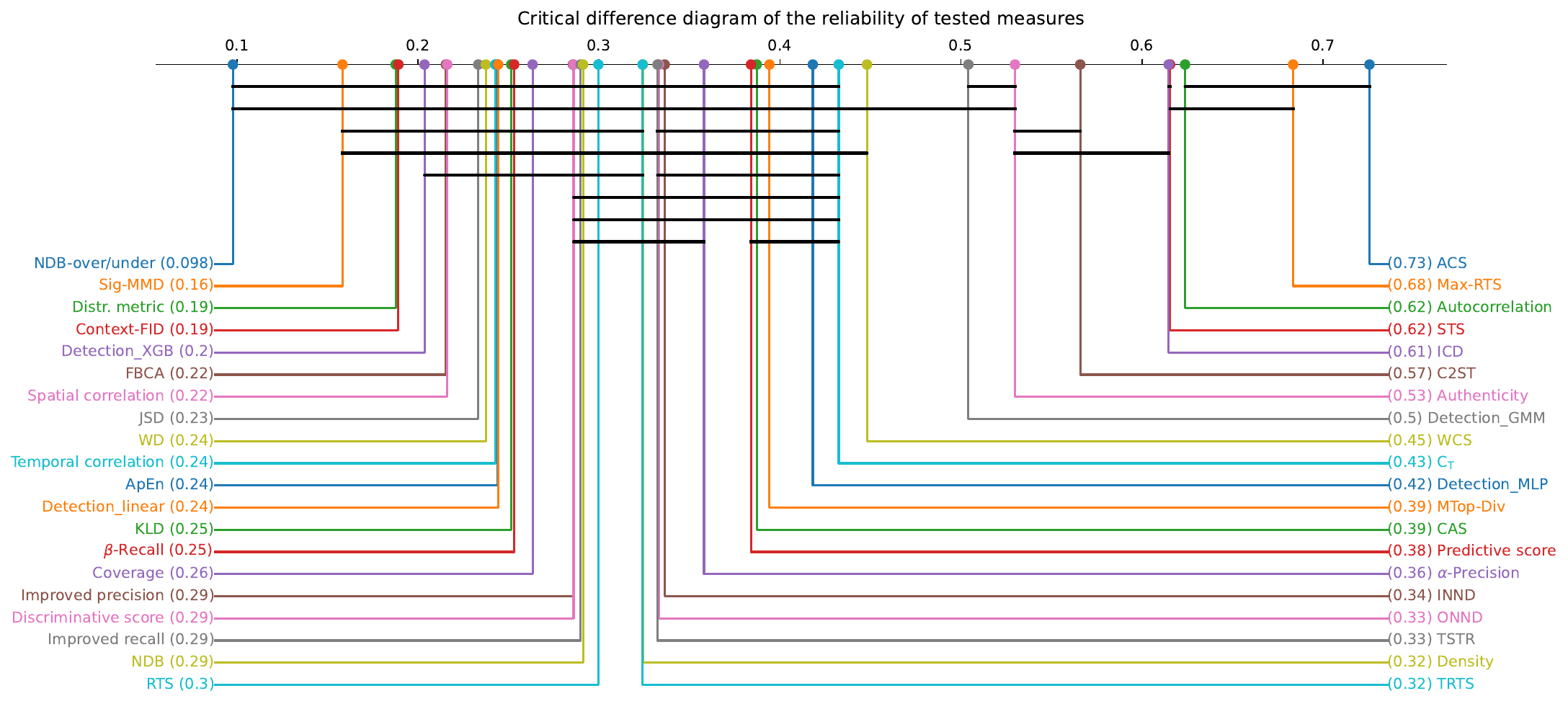}
    \caption{Critical difference diagram for reliability indicator $r_\text{rel}$ in category Generalization as part of \textit{Main}. The horizontal axis at the top depicts $r_\text{rel}$. Additional horizontal bars connect groups of measures with no significantly different $r_\text{rel}$ value.}
    \label{fig:cdd_generalization}
\end{figure}
\end{landscape}

\begin{landscape}
\begin{figure}
    \centering
    \includegraphics[width=\linewidth, trim={0 0 0 8mm}, clip]{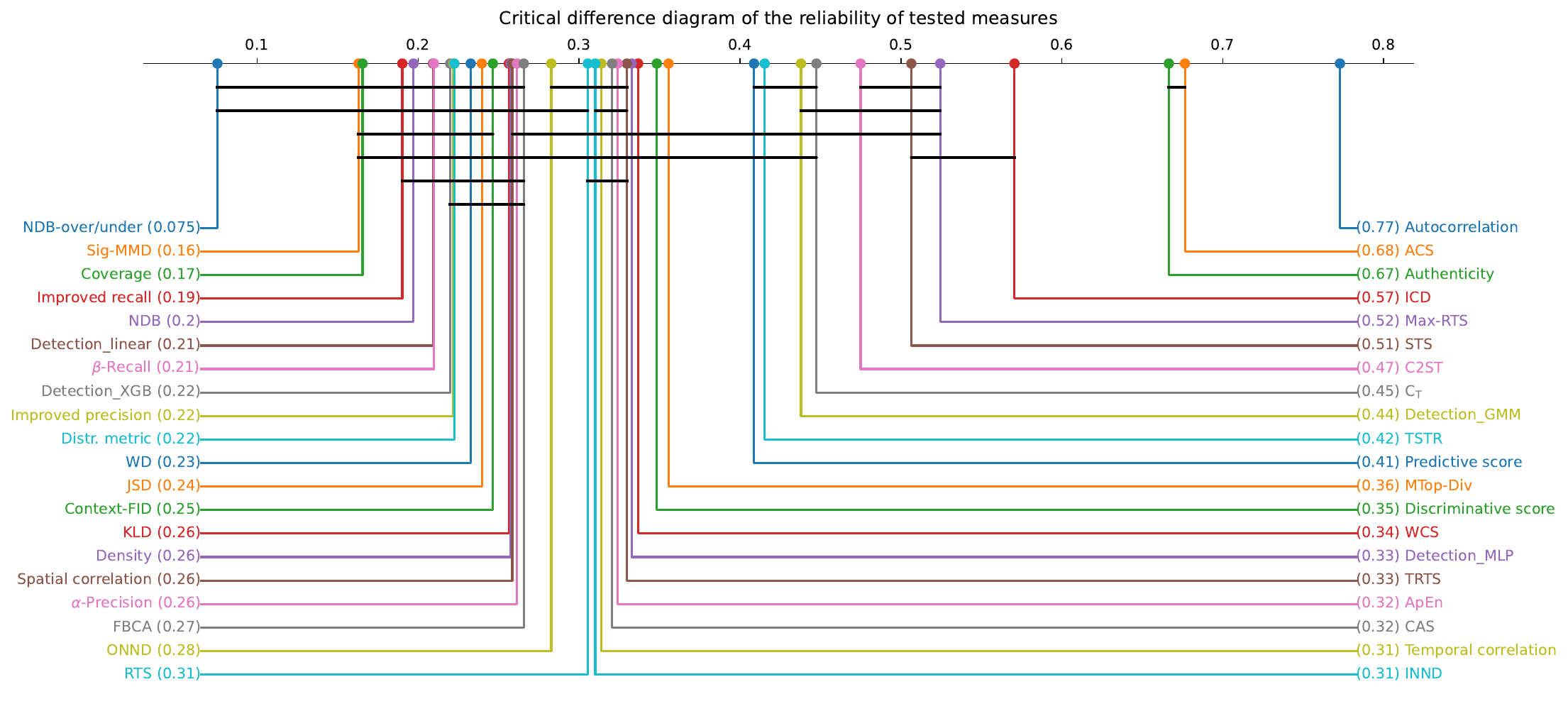}
    \caption{Critical difference diagram for reliability indicator $r_\text{rel}$ in category privacy as part of \textit{Main}. The horizontal axis at the top depicts $r_\text{rel}$. Additional horizontal bars connect groups of measures with no significantly different $r_\text{rel}$ value.}
    \label{fig:cdd_privacy}
\end{figure}
\end{landscape}

\begin{landscape}
\begin{figure}
    \centering
    \includegraphics[width=\linewidth, trim={0 0 0 8mm}, clip]{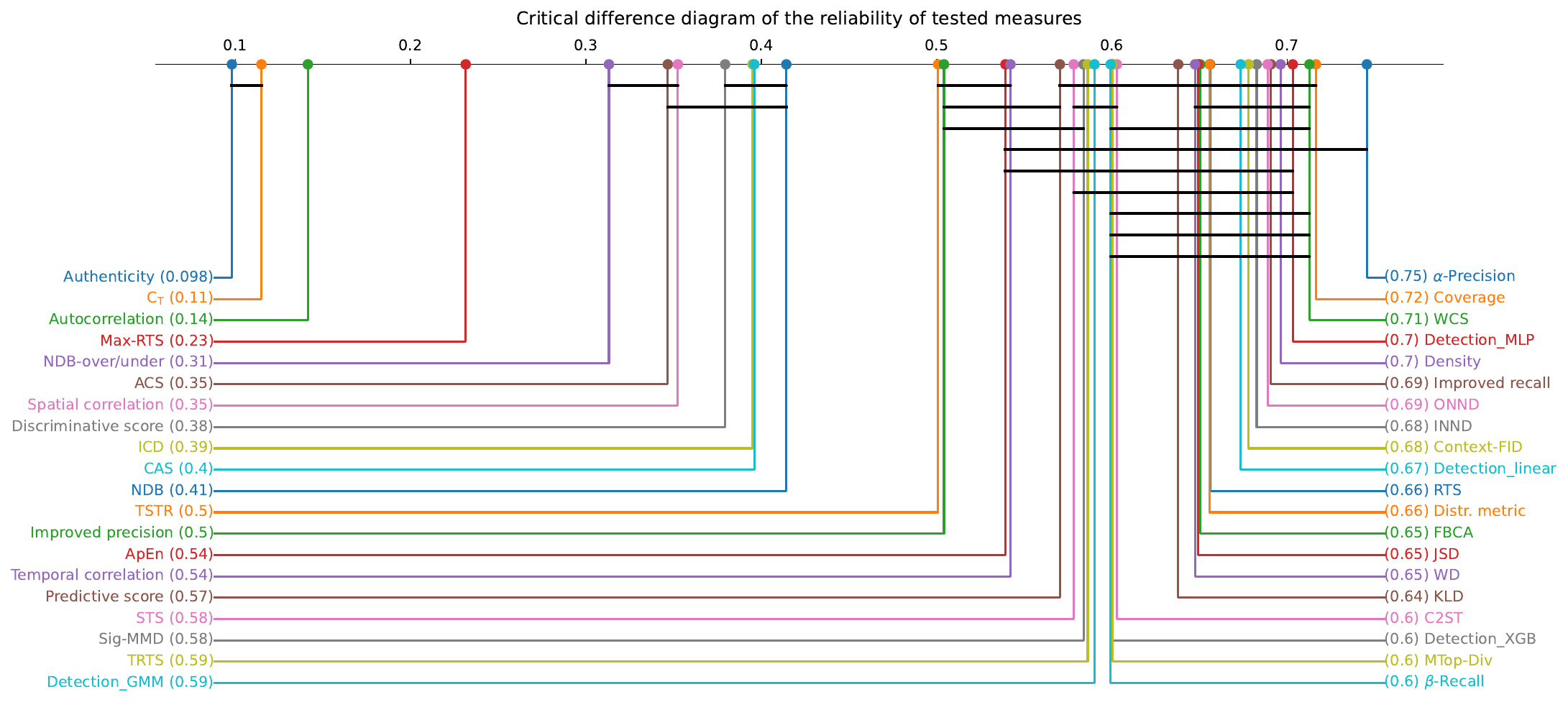}
    \caption{Critical difference diagram for reliability indicator $r_\text{rel}$ in category Representativeness as part of \textit{Main}. The horizontal axis at the top depicts $r_\text{rel}$. Additional horizontal bars connect groups of measures with no significantly different $r_\text{rel}$ value.}
    \label{fig:cdd_representativeness}
\end{figure}
\end{landscape}

Additionally, we used STEB's evaluation component to conducted a statistical analysis of the reliability indicators in each of the four categories. To this end, the Kruskal-Wallis H test~\cite{kruskal-wallis} is employed as omnibus test to determine if there are statistical differences between any of the indicators and the Mann-Whitney U test~\cite{mann-whitney} with Bonferroni correction is applied post-hoc to each pair of measures. The results are visualized in four critical difference diagrams~\cite{cdd}, Figures~\ref{fig:cdd_fidelity} through~\ref{fig:cdd_representativeness}.

\newcommand\hvill{\hspace{0pt plus 1filll}}
\begin{landscape}
\begin{table}[t]
\caption{Measure reliability ranking for experiment \textit{Main}. This is an alternative presentation of \cref{tab:overview}, ranking the measures in each of the four categories by $r_\text{rel}$. For ease of use, mean and standard deviation (Mean $\pm$ StD) are provided with each occurrence of the measure. DOMIAS is again excluded and placed at the bottom.}
\label{tab:ranking}
    \centering
    \scalebox{.78}{
    \begin{tabular}{rcccc}
\toprule
& Fidelity & Generalization & Privacy & Representativeness \\
\midrule
1 & $\alpha$-Precision \hvill $.783 \pm .305$ & ACS \hvill $.726 \pm .320$ & Autocorrelation \hvill $.773 \pm .306$ & $\alpha$-Precision \hvill $.746 \pm .314$ \\
2 & WCS \hvill $.774 \pm .274$ & Max-RTS \hvill $.684 \pm .418$ & ACS \hvill $.676 \pm .313$ & Coverage \hvill $.717 \pm .360$ \\
3 & Coverage \hvill $.770 \pm .323$ & Autocorrelation \hvill $.624 \pm .417$ & Authenticity \hvill $.667 \pm .391$ & WCS \hvill $.713 \pm .278$ \\
4 & Detection\_MLP \hvill $.739 \pm .246$ & STS \hvill $.616 \pm .383$ & ICD \hvill $.571 \pm .303$ & Detection\_MLP \hvill $.703 \pm .236$ \\
5 & Density \hvill $.731 \pm .368$ & ICD \hvill $.615 \pm .314$ & Max-RTS \hvill $.524 \pm .436$ & Density \hvill $.696 \pm .362$ \\
6 & Improved recall \hvill $.715 \pm .339$ & C2ST \hvill $.565 \pm .203$ & STS \hvill $.506 \pm .362$ & Improved recall \hvill $.691 \pm .344$ \\
7 & ONND \hvill $.713 \pm .332$ & Authenticity \hvill $.530 \pm .442$ & C2ST \hvill $.475 \pm .102$ & ONND \hvill $.689 \pm .322$ \\
8 & INND \hvill $.697 \pm .301$ & Detection\_GMM \hvill $.504 \pm .269$ & $\text{C}_\text{T}$ \hvill $.448 \pm .439$ & INND \hvill $.683 \pm .298$ \\
9 & C2ST \hvill $.660 \pm .194$ & WCS \hvill $.448 \pm .329$ & Detection\_GMM \hvill $.438 \pm .200$ & Context-FID \hvill $.678 \pm .406$ \\
10 & Detection\_linear \hvill $.659 \pm .366$ & $\text{C}_\text{T}$ \hvill $.433 \pm .459$ & TSTR \hvill $.415 \pm .184$ & Detection\_linear \hvill $.674 \pm .345$ \\
11 & Detection\_GMM \hvill $.641 \pm .278$ & Detection\_MLP \hvill $.418 \pm .300$ & Predictive score \hvill $.409 \pm .186$ & RTS \hvill $.656 \pm .348$ \\
12 & STS \hvill $.630 \pm .415$ & MTop-Div \hvill $.394 \pm .320$ & MTop-Div \hvill $.355 \pm .299$ & Distr. metric \hvill $.656 \pm .405$ \\
13 & WD \hvill $.617 \pm .404$ & CAS \hvill $.387 \pm .258$ & Discriminative score \hvill $.348 \pm .204$ & FBCA \hvill $.650 \pm .392$ \\
14 & JSD \hvill $.617 \pm .421$ & Predictive score \hvill $.384 \pm .207$ & WCS \hvill $.337 \pm .225$ & JSD \hvill $.649 \pm .393$ \\
15 & $\beta$-Recall \hvill $.608 \pm .435$ & $\alpha$-Precision \hvill $.358 \pm .381$ & Detection\_MLP \hvill $.333 \pm .212$ & WD \hvill $.647 \pm .376$ \\
16 & KLD \hvill $.602 \pm .408$ & INND \hvill $.336 \pm .296$ & TRTS \hvill $.330 \pm .339$ & KLD \hvill $.638 \pm .380$ \\
17 & RTS \hvill $.601 \pm .388$ & ONND \hvill $.333 \pm .320$ & ApEn \hvill $.324 \pm .244$ & C2ST \hvill $.603 \pm .170$ \\
18 & MTop-Div \hvill $.601 \pm .318$ & TSTR \hvill $.332 \pm .228$ & CAS \hvill $.320 \pm .197$ & MTop-Div \hvill $.601 \pm .311$ \\
19 & Context-FID \hvill $.595 \pm .455$ & Density \hvill $.324 \pm .415$ & Temporal correlation \hvill $.314 \pm .229$ & Detection\_XGB \hvill $.600 \pm .401$ \\
20 & Distr. metric \hvill $.594 \pm .438$ & TRTS \hvill $.324 \pm .358$ & INND \hvill $.310 \pm .273$ & $\beta$-Recall \hvill $.599 \pm .425$ \\
21 & FBCA \hvill $.576 \pm .432$ & RTS \hvill $.300 \pm .334$ & RTS \hvill $.305 \pm .307$ & Detection\_GMM \hvill $.590 \pm .257$ \\
22 & Improved precision \hvill $.556 \pm .423$ & NDB \hvill $.291 \pm .343$ & ONND \hvill $.283 \pm .259$ & TRTS \hvill $.586 \pm .397$ \\
23 & TRTS \hvill $.551 \pm .416$ & Improved recall \hvill $.290 \pm .386$ & FBCA \hvill $.266 \pm .355$ & Sig-MMD \hvill $.584 \pm .428$ \\
24 & Predictive score \hvill $.542 \pm .258$ & Discriminative score \hvill $.286 \pm .229$ & $\alpha$-Precision \hvill $.261 \pm .311$ & STS \hvill $.578 \pm .398$ \\
25 & Sig-MMD \hvill $.532 \pm .440$ & Improved precision \hvill $.286 \pm .409$ & Spatial correlation \hvill $.258 \pm .256$ & Predictive score \hvill $.570 \pm .229$ \\
26 & Detection\_XGB \hvill $.530 \pm .424$ & Coverage \hvill $.263 \pm .390$ & Density \hvill $.257 \pm .353$ & Temporal correlation \hvill $.542 \pm .342$ \\
27 & NDB \hvill $.473 \pm .384$ & $\beta$-Recall \hvill $.253 \pm .397$ & KLD \hvill $.256 \pm .319$ & ApEn \hvill $.539 \pm .347$ \\
28 & Temporal correlation \hvill $.462 \pm .382$ & KLD \hvill $.251 \pm .337$ & Context-FID \hvill $.246 \pm .365$ & Improved precision \hvill $.504 \pm .417$ \\
29 & ApEn \hvill $.460 \pm .385$ & Detection\_linear \hvill $.244 \pm .361$ & JSD \hvill $.239 \pm .328$ & TSTR \hvill $.501 \pm .271$ \\
30 & ICD \hvill $.433 \pm .350$ & ApEn \hvill $.244 \pm .247$ & WD \hvill $.233 \pm .302$ & NDB \hvill $.415 \pm .369$ \\
31 & TSTR \hvill $.429 \pm .309$ & Temporal correlation \hvill $.243 \pm .241$ & Distr. metric \hvill $.222 \pm .334$ & CAS \hvill $.396 \pm .259$ \\
32 & ACS \hvill $.416 \pm .413$ & WD \hvill $.237 \pm .330$ & Improved precision \hvill $.222 \pm .354$ & ICD \hvill $.395 \pm .316$ \\
33 & CAS \hvill $.404 \pm .271$ & JSD \hvill $.233 \pm .346$ & Detection\_XGB \hvill $.220 \pm .331$ & Discriminative score \hvill $.379 \pm .252$ \\
34 & Max-RTS \hvill $.382 \pm .458$ & Spatial correlation \hvill $.216 \pm .252$ & $\beta$-Recall \hvill $.210 \pm .349$ & Spatial correlation \hvill $.352 \pm .346$ \\
35 & Discriminative score \hvill $.326 \pm .274$ & FBCA \hvill $.215 \pm .341$ & Detection\_linear \hvill $.209 \pm .328$ & ACS \hvill $.347 \pm .369$ \\
36 & Spatial correlation \hvill $.307 \pm .348$ & Detection\_XGB \hvill $.204 \pm .326$ & NDB \hvill $.197 \pm .255$ & NDB-over/under \hvill $.313 \pm .277$ \\
37 & NDB-over/under \hvill $.295 \pm .280$ & Context-FID \hvill $.189 \pm .348$ & Improved recall \hvill $.190 \pm .308$ & Max-RTS \hvill $.231 \pm .389$ \\
38 & $\text{C}_\text{T}$ \hvill $.158 \pm .314$ & Distr. metric \hvill $.188 \pm .325$ & Coverage \hvill $.165 \pm .311$ & Autocorrelation \hvill $.141 \pm .225$ \\
39 & Autocorrelation \hvill $.083 \pm .165$ & Sig-MMD \hvill $.158 \pm .265$ & Sig-MMD \hvill $.163 \pm .263$ & $\text{C}_\text{T}$ \hvill $.115 \pm .216$ \\
40 & Authenticity \hvill $.052 \pm .124$ & NDB-over/under \hvill $.098 \pm .160$ & NDB-over/under \hvill $.075 \pm .124$ & Authenticity \hvill $.098 \pm .189$ \\
\midrule
- & DOMIAS \hvill $1.   \pm .000$ & DOMIAS \hvill $.000 \pm .000$ & DOMIAS \hvill $.000 \pm .000$ & DOMIAS \hvill $1.   \pm .000$ \\
\bottomrule
    \end{tabular}
    }
\end{table}
\end{landscape}

{\setlength\tabcolsep{5.2pt}
\begin{table}[t]
    \caption{Measure Consistency indicators for experiment \textit{Main}. For each measure, we list $r_\text{con}$ computed for both changing dataset and random seed. The lower $r_\text{con}$, the more the measure scores sway with the choice of parameter. $1.0$ means equal reliability on all datasets/random seeds.}
    \label{tab:consistency}
    \centering
    \vskip 1mm
    {\small
    \rowcolors{2}{gray!25}{white}
    \begin{tabular}{lcccccccc}
        \toprule
         & \multicolumn{2}{c}{Fidelity} & \multicolumn{2}{c}{Generalization} & \multicolumn{2}{c}{Privacy} & \multicolumn{2}{c}{Representativeness} \\
        Measure & Dataset & Seed & Dataset & Seed & Dataset & Seed & Dataset & Seed \\
        \midrule
        $\alpha$-Precision & $.533$ & $1.  $ & $.400$ & $1.  $ & $.400$ & $1.  $ & $.489$ & $1.  $ \\
        $\beta$-Recall & $.422$ & $1.  $ & $.444$ & $1.  $ & $.511$ & $1.  $ & $.444$ & $1.  $ \\
        $\text{C}_\text{T}$ & $.467$ & $1.  $ & $.311$ & $1.  $ & $.311$ & $1.  $ & $.467$ & $1.  $ \\
        ACS & $.133$ & $1.  $ & $.267$ & $1.  $ & $.133$ & $1.  $ & $.111$ & $1.  $ \\
        ApEn & $.244$ & $1.  $ & $.333$ & $1.  $ & $.400$ & $1.  $ & $.400$ & $1.  $ \\
        Authenticity & $.600$ & $1.  $ & $.400$ & $1.  $ & $.444$ & $1.  $ & $.422$ & $1.  $ \\
        Autocorrelation & $.889$ & $1.  $ & $.400$ & $1.  $ & $.356$ & $1.  $ & $.511$ & $1.  $ \\
        C2ST & $.422$ & $1.  $ & $.289$ & $1.  $ & $.444$ & $1.  $ & $.511$ & $1.  $ \\
        CAS & $.200$ & $1.  $ & $.400$ & $1.  $ & $.400$ & $1.  $ & $.400$ & $1.  $ \\
        Context-FID & $.467$ & $1.  $ & $.822$ & $1.  $ & $.578$ & $1.  $ & $.711$ & $1.  $ \\
        Coverage & $.200$ & $1.  $ & $.289$ & $1.  $ & $.444$ & $1.  $ & $.289$ & $1.  $ \\
        Density & $.244$ & $1.  $ & $.178$ & $1.  $ & $.200$ & $1.  $ & $.244$ & $1.  $ \\
        Detection\_GMM & $.422$ & $.689$ & $.622$ & $.533$ & $.800$ & $.533$ & $.311$ & $.689$ \\
        Detection\_MLP & $.333$ & $1.  $ & $.578$ & $1.  $ & $.756$ & $.978$ & $.289$ & $1.  $ \\
        Detection\_XGB & $.444$ & $1.  $ & $.511$ & $1.  $ & $.578$ & $1.  $ & $.422$ & $1.  $ \\
        Detection\_linear & $.156$ & $1.  $ & $.511$ & $1.  $ & $.511$ & $1.  $ & $.178$ & $1.  $ \\
        Discriminative score & $.422$ & $1.  $ & $.333$ & $1.  $ & $.333$ & $.956$ & $.378$ & $.978$ \\
        Distr. metric & $.511$ & $1.  $ & $.733$ & $1.  $ & $.600$ & $1.  $ & $.511$ & $1.  $ \\
        FBCA & $.556$ & $1.  $ & $.778$ & $1.  $ & $.689$ & $1.  $ & $.756$ & $1.  $ \\
        ICD & $.111$ & $1.  $ & $.222$ & $1.  $ & $.267$ & $1.  $ & $.133$ & $1.  $ \\
        INND & $.289$ & $1.  $ & $.444$ & $1.  $ & $.400$ & $1.  $ & $.267$ & $1.  $ \\
        Improved precision & $.133$ & $1.  $ & $.400$ & $1.  $ & $.400$ & $1.  $ & $.156$ & $1.  $ \\
        Improved recall & $.178$ & $1.  $ & $.578$ & $1.  $ & $.600$ & $1.  $ & $.156$ & $1.  $ \\
        JSD & $.844$ & $1.  $ & $.667$ & $1.  $ & $.600$ & $1.  $ & $.622$ & $1.  $ \\
        KLD & $.822$ & $.956$ & $.800$ & $1.  $ & $.933$ & $1.  $ & $.711$ & $1.  $ \\
        MTop-Div & $.200$ & $1.  $ & $.289$ & $1.  $ & $.311$ & $1.  $ & $.222$ & $1.  $ \\
        Max-RTS & $.444$ & $1.  $ & $.511$ & $1.  $ & $.356$ & $1.  $ & $.578$ & $1.  $ \\
        NDB & $.222$ & $1.  $ & $.511$ & $.933$ & $.422$ & $.933$ & $.244$ & $1.  $ \\
        NDB-over/under & $.378$ & $1.  $ & $.467$ & $1.  $ & $.489$ & $1.  $ & $.422$ & $1.  $ \\
        ONND & $.333$ & $1.  $ & $.467$ & $1.  $ & $.533$ & $1.  $ & $.333$ & $1.  $ \\
        Predictive score & $.178$ & $1.  $ & $.511$ & $1.  $ & $.667$ & $1.  $ & $.422$ & $1.  $ \\
        RTS & $.311$ & $1.  $ & $.356$ & $1.  $ & $.489$ & $1.  $ & $.400$ & $1.  $ \\
        STS & $.244$ & $1.  $ & $.311$ & $1.  $ & $.267$ & $1.  $ & $.267$ & $1.  $ \\
        Sig-MMD & $.472$ & $1.  $ & $.417$ & $1.  $ & $.444$ & $1.  $ & $.528$ & $1.  $ \\
        Spatial correlation & $.111$ & $1.  $ & $.244$ & $1.  $ & $.200$ & $1.  $ & $.244$ & $1.  $ \\
        TRTS & $.378$ & $1.  $ & $.333$ & $1.  $ & $.222$ & $1.  $ & $.244$ & $1.  $ \\
        TSTR & $.489$ & $1.  $ & $.578$ & $1.  $ & $.667$ & $1.  $ & $.822$ & $.978$ \\
        Temporal correlation & $.378$ & $1.  $ & $.356$ & $1.  $ & $.556$ & $1.  $ & $.444$ & $1.  $ \\
        WCS & $.306$ & $1.  $ & $.472$ & $1.  $ & $.500$ & $1.  $ & $.361$ & $1.  $ \\
        WD & $.711$ & $.978$ & $.622$ & $1.  $ & $.556$ & $1.  $ & $.556$ & $1.  $ \\
        \midrule
        \rowcolor{white}
        DOMIAS & \multicolumn{8}{c}{N/A} \\
        \bottomrule
    \end{tabular}
}
\end{table}
}

\begin{table}[t]
    \caption{Running time of the measure executions recorded on the \textit{Main} experiment sorted by average rank. The listed values are the average time required to apply the measure to given data across multiple modulation steps and tests. All values are rounded to seconds. Asterisks (*) indicate embedder-dependence, the embedding time is excluded. N/A indicates the absence of successful tests.}
    \label{tab:rt_measures}
    \centering
    \vskip 1mm
    {\small
    \rowcolors{2}{gray!25}{white}
    \begin{tabular}{rlcccccccccc@{\hskip35pt}}
        \toprule
         & Measure & \mcrot{Appliances energy} & \mcrot{ElectricDevices} & \mcrot{Exchange rate} & \mcrot{Google stock} & \mcrot{PPG and respiration} & \mcrot{PTB diagnostic ECG} & \mcrot{Sine} & \mcrot{StarLightCurves} & \mcrot{UniMiB SHAR} & \mcrot{Wikipedia web traffic} \\
        \midrule
        1 & Temporal correlation & $0$ & $0$ & $0$ & $0$ & $0$ & $0$ & $0$ & $0$ & $0$ & $0$ \\
        2 & Spatial correlation & $0$ & $0$ & $0$ & $0$ & $0$ & $0$ & $0$ & $0$ & $0$ & $0$ \\
        3 & Context-FID* & $0$ & $0$ & $0$ & $0$ & $0$ & $0$ & $0$ & $0$ & $0$ & $0$ \\
        4 & FBCA* & $0$ & $0$ & $0$ & $0$ & $0$ & $0$ & $0$ & $0$ & $0$ & $0$ \\
        5 & Improved recall* & $0$ & $0$ & $0$ & $0$ & $0$ & $0$ & $0$ & $0$ & $0$ & $1$ \\
        6 & Improved precision* & $0$ & $0$ & $0$ & $0$ & $0$ & $0$ & $0$ & $0$ & $0$ & $1$ \\
        7 & Detection\_linear* & $0$ & $0$ & $0$ & $0$ & $0$ & $0$ & $0$ & $0$ & $0$ & $1$ \\
        8 & Distr. metric & $0$ & $0$ & $0$ & $0$ & $0$ & $3$ & $0$ & $0$ & $0$ & $0$ \\
        9 & JSD* & $1$ & $0$ & $0$ & $0$ & $0$ & $1$ & $0$ & $0$ & $0$ & $6$ \\
        10 & WD* & $1$ & $1$ & $0$ & $0$ & $0$ & $1$ & $0$ & $0$ & $0$ & $7$ \\
        11 & KLD* & $1$ & $1$ & $0$ & $0$ & $0$ & $2$ & $0$ & $0$ & $0$ & $6$ \\
        12 & ACS & $2$ & $0$ & $0$ & $0$ & $1$ & $28$ & $0$ & $0$ & $0$ & $7$ \\
        13 & NDB-over/under* & $1$ & $0$ & $0$ & $0$ & $1$ & $3$ & $0$ & $0$ & $1$ & $4$ \\
        14 & Sig-MMD & $1$ & $1$ & $0$ & $0$ & $1$ & N/A & $0$ & $1$ & $0$ & $2$ \\
        15 & NDB* & $1$ & $1$ & $0$ & $0$ & $1$ & $2$ & $0$ & $0$ & $0$ & $9$ \\
        16 & Autocorrelation & $5$ & $0$ & $0$ & $0$ & $1$ & $660$ & $0$ & $1$ & $0$ & $12$ \\
        17 & $\text{C}_\text{T}$* & $1$ & $1$ & $0$ & $0$ & $1$ & $3$ & $0$ & $0$ & $1$ & $6$ \\
        18 & Density* & $1$ & $2$ & $0$ & $0$ & $2$ & $3$ & $1$ & $0$ & $1$ & $3$ \\
        19 & ApEn & $4$ & $0$ & $1$ & $1$ & $1$ & $16$ & $0$ & $1$ & $0$ & $0$ \\
        20 & Coverage* & $1$ & $1$ & $0$ & $0$ & $2$ & $3$ & $1$ & $1$ & $1$ & $3$ \\
        21 & Detection\_XGB* & $1$ & $2$ & $1$ & $1$ & $2$ & $4$ & $1$ & $1$ & $2$ & $6$ \\
        22 & Max-RTS* & $3$ & $3$ & $1$ & $0$ & $3$ & $12$ & $1$ & $1$ & $2$ & $30$ \\
        23 & STS* & $4$ & $3$ & $1$ & $1$ & $4$ & $11$ & $2$ & $2$ & $2$ & $21$ \\
        24 & ICD & $3$ & $3$ & $3$ & $3$ & $3$ & $11$ & $3$ & $12$ & $3$ & $4$ \\
        25 & INND & $3$ & $3$ & $3$ & $3$ & $3$ & $12$ & $3$ & $11$ & $3$ & $4$ \\
        26 & ONND & $3$ & $3$ & $3$ & $3$ & $3$ & $11$ & $3$ & $11$ & $3$ & $4$ \\
        27 & RTS* & $5$ & $4$ & $2$ & $1$ & $5$ & $18$ & $2$ & $2$ & $3$ & $49$ \\
        28 & Detection\_GMM* & $8$ & $11$ & $2$ & $1$ & $17$ & $49$ & $4$ & $3$ & $3$ & $71$ \\
        29 & $\beta$-Recall* & $9$ & $8$ & $4$ & $2$ & $15$ & $33$ & $6$ & $5$ & $6$ & $67$ \\
        30 & $\alpha$-Precision* & $9$ & $9$ & $3$ & $2$ & $14$ & $40$ & $5$ & $5$ & $7$ & $70$ \\
        31 & Authenticity* & $8$ & $10$ & $4$ & $2$ & $11$ & $42$ & $5$ & $4$ & $6$ & $77$ \\
        32 & TRTS & $25$ & $11$ & $9$ & $4$ & $26$ & $59$ & $12$ & $14$ & $15$ & $88$ \\
        33 & Predictive score & $26$ & $10$ & $11$ & $5$ & $24$ & $63$ & $11$ & $13$ & $16$ & $87$ \\
        34 & Discriminative score & $25$ & $23$ & $9$ & $5$ & $36$ & $124$ & $16$ & $22$ & $19$ & $216$ \\
        35 & C2ST & $22$ & $23$ & $10$ & $5$ & $40$ & $125$ & $16$ & $22$ & $20$ & $219$ \\
        36 & Detection\_MLP* & $34$ & $32$ & $12$ & $6$ & $43$ & $85$ & $18$ & $17$ & $20$ & $157$ \\
        37 & MTop-Div* & $100$ & $38$ & $54$ & $40$ & $40$ & $42$ & $38$ & $42$ & $40$ & $40$ \\
        38 & TSTR & $46$ & $20$ & $19$ & $8$ & $51$ & $122$ & $23$ & $25$ & $31$ & $168$ \\
        39 & WCS & $430$ & $9$ & $26$ & $18$ & $54$ & N/A & $19$ & $130$ & $48$ & $77$ \\
        40 & CAS & N/A & $80$ & N/A & N/A & N/A & $466$ & $30$ & $41$ & $90$ & N/A \\
        41 & DOMIAS* & \multicolumn{10}{c}{N/A} \\
\bottomrule
\end{tabular}
}
\end{table}

\begin{landscape}
{\setlength\tabcolsep{3.7pt}
\begin{table}[t]
    \caption{Running time statistics of the measure executions extending \cref{tab:rt_measures} for \emph{Appliances energy}, \emph{ElectricDevices}, \emph{Exchange rate}, \emph{Google stock}, and \emph{PPG and respiration}. In addition to the average running time (Mean), we provide for each dataset and measure the standard deviation (StD) of the measurements, the number of complete, un-aided executions (Valid), and the cache-aided executions (Cached). Aided means that the execution was accelerated by the use of previously computed and cached artifacts. Running times for DOMIAS not available.}
    \label{tab:rt_measures_long1}
    \centering
    \scalebox{.75}{
    \rowcolors{2}{gray!25}{white}
    \begin{tabular}{@{\hskip0pt}rl *{20}c@{\hskip0pt}}
        \toprule
         & Measure & \multicolumn{4}{c}{Appliances energy} & \multicolumn{4}{c}{ElectricDevices} & \multicolumn{4}{c}{Exchange rate} & \multicolumn{4}{c}{Google stock} & \multicolumn{4}{c}{PPG and respiration} \\
         &  & Mean & StD & Valid & Cached & Mean & StD & Valid & Cached & Mean & StD & Valid & Cached & Mean & StD & Valid & Cached & Mean & StD & Valid & Cached \\
        \midrule
        1 & Temporal correlation & $0$ & $0$ & $20$ & $970$ & $0$ & $0$ & $20$ & $1190$ & $0$ & $0$ & $20$ & $970$ & $0$ & $0$ & $20$ & $970$ & $0$ & $0$ & $20$ & $970$ \\
        2 & Spatial correlation & $0$ & $0$ & $20$ & $970$ & $0$ & $0$ & $20$ & $1190$ & $0$ & $0$ & $20$ & $970$ & $0$ & $0$ & $20$ & $948$ & $0$ & $0$ & $20$ & $970$ \\
        3 & Improved recall & $0$ & $0$ & $990$ & $0$ & $0$ & $0$ & $1210$ & $0$ & $0$ & $0$ & $990$ & $0$ & $0$ & $0$ & $990$ & $0$ & $0$ & $0$ & $990$ & $0$ \\
        4 & Context-FID & $0$ & $0$ & $990$ & $0$ & $0$ & $0$ & $1210$ & $0$ & $0$ & $0$ & $990$ & $0$ & $0$ & $0$ & $990$ & $0$ & $0$ & $0$ & $990$ & $0$ \\
        5 & FBCA & $0$ & $0$ & $20$ & $970$ & $0$ & $0$ & $20$ & $1190$ & $0$ & $0$ & $20$ & $970$ & $0$ & $0$ & $20$ & $970$ & $0$ & $0$ & $20$ & $970$ \\
        6 & Improved precision & $0$ & $0$ & $20$ & $970$ & $0$ & $0$ & $20$ & $1190$ & $0$ & $0$ & $20$ & $970$ & $0$ & $0$ & $20$ & $970$ & $0$ & $0$ & $20$ & $970$ \\
        7 & Distr. metric & $0$ & $0$ & $20$ & $970$ & $0$ & $0$ & $20$ & $1190$ & $0$ & $0$ & $20$ & $970$ & $0$ & $0$ & $20$ & $970$ & $0$ & $0$ & $20$ & $970$ \\
        8 & Detection\_linear & $0$ & $0$ & $990$ & $0$ & $0$ & $0$ & $1210$ & $0$ & $0$ & $0$ & $990$ & $0$ & $0$ & $0$ & $990$ & $0$ & $0$ & $0$ & $990$ & $0$ \\
        9 & JSD & $1$ & $0$ & $20$ & $959$ & $0$ & $0$ & $20$ & $1190$ & $0$ & $0$ & $20$ & $970$ & $0$ & $0$ & $20$ & $970$ & $0$ & $0$ & $20$ & $970$ \\
        10 & ACS & $2$ & $0$ & $990$ & $0$ & $0$ & $0$ & $1210$ & $0$ & $0$ & $0$ & $990$ & $0$ & $0$ & $0$ & $990$ & $0$ & $1$ & $0$ & $990$ & $0$ \\
        11 & Autocorrelation & $5$ & $2$ & $20$ & $970$ & $0$ & $0$ & $20$ & $1190$ & $0$ & $0$ & $20$ & $970$ & $0$ & $0$ & $20$ & $970$ & $1$ & $0$ & $20$ & $970$ \\
        12 & WD & $1$ & $0$ & $20$ & $948$ & $1$ & $0$ & $20$ & $1190$ & $0$ & $0$ & $20$ & $970$ & $0$ & $0$ & $20$ & $970$ & $0$ & $0$ & $20$ & $970$ \\
        13 & KLD & $1$ & $0$ & $20$ & $970$ & $1$ & $0$ & $20$ & $1190$ & $0$ & $0$ & $20$ & $970$ & $0$ & $0$ & $20$ & $970$ & $0$ & $0$ & $20$ & $970$ \\
        14 & NDB & $1$ & $0$ & $20$ & $970$ & $1$ & $0$ & $20$ & $1190$ & $0$ & $0$ & $20$ & $970$ & $0$ & $0$ & $20$ & $970$ & $1$ & $0$ & $20$ & $970$ \\
        15 & Sig-MMD & $1$ & $0$ & $737$ & $0$ & $1$ & $0$ & $1210$ & $0$ & $0$ & $0$ & $990$ & $0$ & $0$ & $0$ & $990$ & $0$ & $1$ & $0$ & $473$ & $0$ \\
        16 & NDB-over/under & $1$ & $0$ & $20$ & $849$ & $0$ & $0$ & $20$ & $1190$ & $0$ & $0$ & $20$ & $871$ & $0$ & $0$ & $20$ & $849$ & $1$ & $0$ & $20$ & $948$ \\
        17 & $\text{C}_\text{T}$ & $1$ & $0$ & $20$ & $937$ & $1$ & $0$ & $20$ & $1190$ & $0$ & $0$ & $20$ & $970$ & $0$ & $0$ & $20$ & $904$ & $1$ & $0$ & $20$ & $970$ \\
        18 & Density & $1$ & $0$ & $10$ & $980$ & $2$ & $1$ & $8$ & $1202$ & $0$ & $0$ & $11$ & $979$ & $0$ & $0$ & $9$ & $981$ & $2$ & $1$ & $11$ & $979$ \\
        19 & Coverage & $1$ & $0$ & $10$ & $980$ & $1$ & $1$ & $12$ & $1198$ & $0$ & $0$ & $9$ & $981$ & $0$ & $0$ & $11$ & $979$ & $2$ & $1$ & $9$ & $981$ \\
        20 & ApEn & $4$ & $0$ & $20$ & $970$ & $0$ & $0$ & $20$ & $1190$ & $1$ & $0$ & $20$ & $970$ & $1$ & $0$ & $20$ & $970$ & $1$ & $0$ & $20$ & $970$ \\
        21 & Detection\_XGB & $1$ & $1$ & $990$ & $0$ & $2$ & $0$ & $1210$ & $0$ & $1$ & $0$ & $990$ & $0$ & $1$ & $0$ & $990$ & $0$ & $2$ & $0$ & $990$ & $0$ \\
        22 & Max-RTS & $3$ & $4$ & $990$ & $0$ & $3$ & $3$ & $1210$ & $0$ & $1$ & $1$ & $990$ & $0$ & $0$ & $1$ & $990$ & $0$ & $3$ & $4$ & $990$ & $0$ \\
        23 & STS & $4$ & $1$ & $968$ & $0$ & $3$ & $1$ & $1210$ & $0$ & $1$ & $0$ & $990$ & $0$ & $1$ & $0$ & $990$ & $0$ & $4$ & $1$ & $990$ & $0$ \\
        24 & ICD & $3$ & $0$ & $990$ & $0$ & $3$ & $0$ & $1210$ & $0$ & $3$ & $0$ & $990$ & $0$ & $3$ & $0$ & $990$ & $0$ & $3$ & $0$ & $990$ & $0$ \\
        25 & INND & $3$ & $0$ & $990$ & $0$ & $3$ & $0$ & $1210$ & $0$ & $3$ & $0$ & $990$ & $0$ & $3$ & $0$ & $990$ & $0$ & $3$ & $0$ & $990$ & $0$ \\
        26 & RTS & $5$ & $1$ & $20$ & $970$ & $4$ & $1$ & $20$ & $1190$ & $2$ & $0$ & $20$ & $970$ & $1$ & $0$ & $20$ & $970$ & $5$ & $1$ & $20$ & $959$ \\
        27 & ONND & $3$ & $0$ & $990$ & $0$ & $3$ & $1$ & $1210$ & $0$ & $3$ & $0$ & $990$ & $0$ & $3$ & $1$ & $990$ & $0$ & $3$ & $1$ & $990$ & $0$ \\
        28 & Detection\_GMM & $8$ & $7$ & $990$ & $0$ & $11$ & $7$ & $1199$ & $0$ & $2$ & $2$ & $990$ & $0$ & $1$ & $1$ & $990$ & $0$ & $17$ & $10$ & $990$ & $0$ \\
        29 & $\alpha$-Precision & $9$ & $2$ & $8$ & $982$ & $9$ & $2$ & $6$ & $1204$ & $3$ & $1$ & $6$ & $984$ & $2$ & $0$ & $9$ & $981$ & $14$ & $6$ & $4$ & $986$ \\
        30 & $\beta$-Recall & $9$ & $2$ & $5$ & $985$ & $8$ & $2$ & $4$ & $1206$ & $4$ & $1$ & $8$ & $982$ & $2$ & $0$ & $6$ & $984$ & $15$ & $5$ & $8$ & $982$ \\
        31 & Authenticity & $8$ & $2$ & $7$ & $972$ & $10$ & $2$ & $10$ & $1200$ & $4$ & $1$ & $6$ & $984$ & $2$ & $0$ & $5$ & $985$ & $11$ & $2$ & $8$ & $982$ \\
        32 & TRTS & $25$ & $5$ & $20$ & $970$ & $11$ & $2$ & $20$ & $1190$ & $9$ & $2$ & $20$ & $970$ & $4$ & $1$ & $20$ & $970$ & $26$ & $7$ & $20$ & $970$ \\
        33 & Predictive score & $26$ & $11$ & $979$ & $0$ & $10$ & $2$ & $1210$ & $0$ & $11$ & $4$ & $990$ & $0$ & $5$ & $3$ & $990$ & $0$ & $24$ & $6$ & $990$ & $0$ \\
        34 & C2ST & $22$ & $6$ & $979$ & $0$ & $23$ & $14$ & $1210$ & $0$ & $10$ & $6$ & $990$ & $0$ & $5$ & $2$ & $990$ & $0$ & $40$ & $28$ & $990$ & $0$ \\
        35 & Discriminative score & $25$ & $9$ & $990$ & $0$ & $23$ & $12$ & $1210$ & $0$ & $9$ & $5$ & $979$ & $0$ & $5$ & $2$ & $990$ & $0$ & $36$ & $27$ & $990$ & $0$ \\
        36 & Detection\_MLP & $34$ & $23$ & $990$ & $0$ & $32$ & $21$ & $1199$ & $0$ & $12$ & $7$ & $990$ & $0$ & $6$ & $3$ & $990$ & $0$ & $43$ & $27$ & $979$ & $0$ \\
        37 & TSTR & $46$ & $9$ & $20$ & $970$ & $20$ & $4$ & $20$ & $1190$ & $19$ & $4$ & $20$ & $970$ & $8$ & $1$ & $20$ & $970$ & $51$ & $10$ & $20$ & $970$ \\
        38 & WCS & $430$ & $5$ & $880$ & $0$ & $9$ & $2$ & $1210$ & $0$ & $26$ & $2$ & $990$ & $0$ & $18$ & $2$ & $990$ & $0$ & $54$ & $6$ & $990$ & $0$ \\
        39 & MTop-Div & $100$ & $90$ & $990$ & $0$ & $38$ & $7$ & $1210$ & $0$ & $54$ & $62$ & $990$ & $0$ & $40$ & $11$ & $990$ & $0$ & $40$ & $7$ & $990$ & $0$ \\
        40 & CAS & NaN & NaN & NaN & NaN & $80$ & $27$ & $20$ & $1300$ & NaN & NaN & NaN & NaN & NaN & NaN & NaN & NaN & NaN & NaN & NaN & NaN \\
\bottomrule
\end{tabular}
}
\end{table}
}
\end{landscape}

\begin{landscape}
{\setlength\tabcolsep{3.9pt}
\begin{table}[t]
    \caption{Running time statistics of the measure executions extending \cref{tab:rt_measures} for \emph{PTB diagnostic ECG}, \emph{Sine}, \emph{StarLightCurves}, \emph{UniMiB SHAR}, and \emph{Wikipedia web traffic}. In addition to the average running time (Mean), we provide for each dataset and measure the standard deviation (StD) of the measurements, the number of complete, un-aided executions (Valid), and the cache-aided executions (Cached). Aided means that the execution was accelerated by the use of previously computed and cached artifacts. Running times for DOMIAS not available.}
    \label{tab:rt_measures_long2}
    \centering
    \scalebox{.75}{
    \rowcolors{2}{gray!25}{white}
    \begin{tabular}{@{\hskip0pt}rl *{20}c@{\hskip0pt}}
        \toprule
         & Measure & \multicolumn{4}{c}{PTB diagnostic ECG} & \multicolumn{4}{c}{Sine} & \multicolumn{4}{c}{StarLightCurves} & \multicolumn{4}{c}{UniMiB SHAR} & \multicolumn{4}{c}{Wikipedia web traffic} \\
         &  & Mean & StD & Valid & Cached & Mean & StD & Valid & Cached & Mean & StD & Valid & Cached & Mean & StD & Valid & Cached & Mean & StD & Valid & Cached \\
        \midrule
        1 & Spatial correlation & $0$ & $0$ & $19$ & $1191$ & $0$ & $0$ & $20$ & $1300$ & $0$ & $0$ & $20$ & $1190$ & $0$ & $0$ & $20$ & $1300$ & $0$ & $0$ & $19$ & $751$ \\
        2 & Temporal correlation & $0$ & $0$ & $35$ & $1175$ & $0$ & $0$ & $20$ & $1300$ & $0$ & $0$ & $20$ & $1190$ & $0$ & $0$ & $20$ & $1300$ & $0$ & $0$ & $16$ & $754$ \\
        3 & Context-FID & $0$ & $0$ & $1210$ & $0$ & $0$ & $0$ & $1320$ & $0$ & $0$ & $0$ & $1199$ & $0$ & $0$ & $0$ & $1320$ & $0$ & $0$ & $0$ & $770$ & $0$ \\
        4 & Improved recall & $0$ & $0$ & $1210$ & $0$ & $0$ & $0$ & $1320$ & $0$ & $0$ & $0$ & $1210$ & $0$ & $0$ & $0$ & $1320$ & $0$ & $1$ & $0$ & $759$ & $0$ \\
        5 & FBCA & $0$ & $0$ & $25$ & $1185$ & $0$ & $0$ & $20$ & $1300$ & $0$ & $0$ & $20$ & $1190$ & $0$ & $0$ & $20$ & $1300$ & $0$ & $0$ & $16$ & $754$ \\
        6 & Improved precision & $0$ & $0$ & $20$ & $1190$ & $0$ & $0$ & $20$ & $1300$ & $0$ & $0$ & $20$ & $1179$ & $0$ & $0$ & $20$ & $1300$ & $1$ & $0$ & $20$ & $750$ \\
        7 & Detection\_linear & $0$ & $0$ & $1188$ & $0$ & $0$ & $0$ & $1320$ & $0$ & $0$ & $0$ & $1210$ & $0$ & $0$ & $0$ & $1320$ & $0$ & $1$ & $1$ & $770$ & $0$ \\
        8 & Distr. metric & $3$ & $1$ & $17$ & $1182$ & $0$ & $0$ & $20$ & $1300$ & $0$ & $0$ & $20$ & $1190$ & $0$ & $0$ & $20$ & $1300$ & $0$ & $0$ & $32$ & $738$ \\
        9 & JSD & $1$ & $1$ & $19$ & $1191$ & $0$ & $0$ & $20$ & $1300$ & $0$ & $0$ & $20$ & $1190$ & $0$ & $0$ & $20$ & $1300$ & $6$ & $2$ & $18$ & $752$ \\
        10 & WD & $1$ & $0$ & $23$ & $1187$ & $0$ & $0$ & $20$ & $1300$ & $0$ & $0$ & $20$ & $1190$ & $0$ & $0$ & $20$ & $1300$ & $7$ & $6$ & $17$ & $742$ \\
        11 & KLD & $2$ & $1$ & $23$ & $1176$ & $0$ & $0$ & $20$ & $1300$ & $0$ & $0$ & $20$ & $1179$ & $0$ & $0$ & $20$ & $1300$ & $6$ & $3$ & $17$ & $731$ \\
        12 & NDB-over/under & $3$ & $1$ & $20$ & $750$ & $0$ & $0$ & $20$ & $1300$ & $0$ & $0$ & $20$ & $1179$ & $1$ & $0$ & $20$ & $1300$ & $4$ & $1$ & $18$ & $730$ \\
        13 & ApEn & $16$ & $7$ & $20$ & $1190$ & $0$ & $0$ & $20$ & $1300$ & $1$ & $0$ & $20$ & $1190$ & $0$ & $0$ & $20$ & $1300$ & $0$ & $0$ & $20$ & $750$ \\
        14 & Sig-MMD & N/A & N/A & N/A & N/A & $0$ & $0$ & $1320$ & $0$ & $1$ & $0$ & $330$ & $0$ & $0$ & $0$ & $1320$ & $0$ & $2$ & $0$ & $231$ & $0$ \\
        15 & Density & $3$ & $0$ & $9$ & $1201$ & $1$ & $0$ & $13$ & $1307$ & $0$ & $0$ & $10$ & $1189$ & $1$ & $0$ & $13$ & $1307$ & $3$ & $0$ & $14$ & $756$ \\
        16 & NDB & $2$ & $1$ & $17$ & $1171$ & $0$ & $0$ & $20$ & $1300$ & $0$ & $0$ & $20$ & $1168$ & $0$ & $0$ & $20$ & $1300$ & $9$ & $4$ & $18$ & $697$ \\
        17 & Coverage & $3$ & $0$ & $11$ & $1199$ & $1$ & $0$ & $7$ & $1313$ & $1$ & $0$ & $10$ & $1200$ & $1$ & $0$ & $7$ & $1313$ & $3$ & $0$ & $20$ & $750$ \\
        18 & ACS & $28$ & $26$ & $1210$ & $0$ & $0$ & $0$ & $1320$ & $0$ & $0$ & $0$ & $1210$ & $0$ & $0$ & $0$ & $1320$ & $0$ & $7$ & $30$ & $770$ & $0$ \\
        19 & $\text{C}_\text{T}$ & $3$ & $1$ & $18$ & $1181$ & $0$ & $0$ & $20$ & $1300$ & $0$ & $0$ & $20$ & $1179$ & $1$ & $0$ & $20$ & $1300$ & $6$ & $2$ & $19$ & $740$ \\
        20 & Autocorrelation & $660$ & $165$ & $37$ & $1173$ & $0$ & $0$ & $20$ & $1300$ & $1$ & $1$ & $20$ & $1190$ & $0$ & $0$ & $20$ & $1300$ & $12$ & $6$ & $19$ & $751$ \\
        21 & Detection\_XGB & $4$ & $1$ & $1210$ & $0$ & $1$ & $0$ & $1320$ & $0$ & $1$ & $0$ & $1210$ & $0$ & $2$ & $0$ & $1320$ & $0$ & $6$ & $2$ & $770$ & $0$ \\
        22 & Max-RTS & $12$ & $11$ & $1210$ & $0$ & $1$ & $1$ & $1320$ & $0$ & $1$ & $1$ & $1210$ & $0$ & $2$ & $1$ & $1320$ & $0$ & $30$ & $20$ & $759$ & $0$ \\
        23 & STS & $11$ & $2$ & $1210$ & $0$ & $2$ & $0$ & $1320$ & $0$ & $2$ & $0$ & $1210$ & $0$ & $2$ & $0$ & $1320$ & $0$ & $21$ & $4$ & $770$ & $0$ \\
        24 & ICD & $11$ & $7$ & $1210$ & $0$ & $3$ & $0$ & $1320$ & $0$ & $12$ & $10$ & $1210$ & $0$ & $3$ & $1$ & $1320$ & $0$ & $4$ & $0$ & $770$ & $0$ \\
        25 & ONND & $11$ & $8$ & $1210$ & $0$ & $3$ & $0$ & $1320$ & $0$ & $11$ & $10$ & $1210$ & $0$ & $3$ & $1$ & $1320$ & $0$ & $4$ & $0$ & $770$ & $0$ \\
        26 & INND & $12$ & $9$ & $1210$ & $0$ & $3$ & $0$ & $1320$ & $0$ & $11$ & $10$ & $1210$ & $0$ & $3$ & $0$ & $1320$ & $0$ & $4$ & $0$ & $770$ & $0$ \\
        27 & RTS & $18$ & $5$ & $20$ & $1190$ & $2$ & $0$ & $20$ & $1300$ & $2$ & $0$ & $20$ & $1190$ & $3$ & $1$ & $20$ & $1300$ & $49$ & $19$ & $18$ & $752$ \\
        28 & Detection\_GMM & $49$ & $25$ & $1210$ & $0$ & $4$ & $3$ & $1320$ & $0$ & $3$ & $1$ & $1199$ & $0$ & $3$ & $1$ & $1320$ & $0$ & $71$ & $36$ & $770$ & $0$ \\
        29 & $\beta$-Recall & $33$ & $9$ & $8$ & $1202$ & $6$ & $1$ & $7$ & $1313$ & $5$ & $1$ & $7$ & $1203$ & $6$ & $1$ & $9$ & $1311$ & $67$ & $19$ & $7$ & $763$ \\
        30 & Authenticity & $42$ & $16$ & $6$ & $1204$ & $5$ & $1$ & $5$ & $1315$ & $4$ & $1$ & $4$ & $1195$ & $6$ & $2$ & $5$ & $1315$ & $77$ & $26$ & $7$ & $763$ \\
        31 & $\alpha$-Precision & $40$ & $7$ & $6$ & $1193$ & $5$ & $1$ & $8$ & $1312$ & $5$ & $1$ & $9$ & $1190$ & $7$ & $1$ & $6$ & $1314$ & $70$ & $16$ & $6$ & $764$ \\
        32 & Predictive score & $63$ & $18$ & $1210$ & $0$ & $11$ & $4$ & $1320$ & $0$ & $13$ & $5$ & $1210$ & $0$ & $16$ & $6$ & $1320$ & $0$ & $87$ & $19$ & $770$ & $0$ \\
        33 & TRTS & $59$ & $14$ & $14$ & $1119$ & $12$ & $3$ & $20$ & $1300$ & $14$ & $4$ & $20$ & $1179$ & $15$ & $4$ & $20$ & $1300$ & $88$ & $19$ & $17$ & $753$ \\
        34 & MTop-Div & $42$ & $6$ & $1210$ & $0$ & $38$ & $14$ & $1320$ & $0$ & $42$ & $7$ & $1210$ & $0$ & $40$ & $6$ & $1320$ & $0$ & $40$ & $5$ & $770$ & $0$ \\
        35 & Detection\_MLP & $85$ & $45$ & $1210$ & $0$ & $18$ & $11$ & $1320$ & $0$ & $17$ & $12$ & $1144$ & $66$ & $20$ & $14$ & $1320$ & $0$ & $157$ & $107$ & $748$ & $0$ \\
        36 & Discriminative score & $124$ & $71$ & $1199$ & $0$ & $16$ & $9$ & $1320$ & $0$ & $22$ & $11$ & $1210$ & $0$ & $19$ & $11$ & $1320$ & $0$ & $216$ & $129$ & $748$ & $0$ \\
        37 & C2ST & $125$ & $72$ & $1199$ & $0$ & $16$ & $9$ & $1320$ & $0$ & $22$ & $11$ & $1199$ & $0$ & $20$ & $13$ & $1320$ & $0$ & $219$ & $126$ & $759$ & $0$ \\
        38 & TSTR & $122$ & $27$ & $17$ & $1193$ & $23$ & $6$ & $20$ & $1289$ & $25$ & $8$ & $20$ & $1190$ & $31$ & $9$ & $20$ & $1300$ & $168$ & $38$ & $19$ & $740$ \\
        39 & WCS & N/A & N/A & N/A & N/A & $19$ & $3$ & $1320$ & $0$ & $130$ & $7$ & $1210$ & $0$ & $48$ & $3$ & $1320$ & $0$ & $77$ & $3$ & $770$ & $0$ \\
        40 & CAS & $466$ & $163$ & $17$ & $1292$ & $30$ & $5$ & $20$ & $1410$ & $41$ & $16$ & $20$ & $1300$ & $90$ & $22$ & $20$ & $1410$ & N/A & N/A & N/A & N/A \\
\bottomrule
\end{tabular}
}
\end{table}
}
\end{landscape}

\begin{table}[t]
    \caption{Running time of the embedder usage recorded on the Main and Embedders experiments sorted by average rank. The listed values are the average time required to employ the embedder across multiple tests, combining inference and training. All values are rounded to seconds.}
    \label{tab:rt_embedders}
    \centering
\begin{tabular}{rl *{10}c@{\hskip39pt}}
\toprule
 & Embedding & \mcrot{Appliances energy} & \mcrot{ElectricDevices} & \mcrot{Exchange rate} & \mcrot{Google stock} & \mcrot{PPG and respiration} & \mcrot{PTB diagnostic ECG} & \mcrot{Sine} & \mcrot{StarLightCurves} & \mcrot{UniMiB SHAR} & \mcrot{Wikipedia web traffic} \\
\midrule
1 & Concat & $0$ & $0$ & $0$ & $0$ & $0$ & $0$ & $0$ & $0$ & $0$ & $0$ \\
2 & Catch22 & $99$ & $2$ & $3$ & $1$ & $14$ & $889$ & $3$ & $12$ & $7$ & $86$ \\
3 & TS2Vec & $327$ & $391$ & $41$ & $38$ & $515$ & $3697$ & $252$ & $640$ & $256$ & $4812$ \\
\bottomrule
\end{tabular}
\end{table}

\begin{table}[t]
    \caption{Running time statistics of the embedder usage extending \cref{tab:rt_embedders}. In addition to the average running time (Mean), we provide for each dataset and measure the standard deviation (StD) of the measurements, the number of complete, the number of completed, un-aided executions (Valid), and the cache-aided executions (Cached). Aided means that the model training was skipped due to loading a cached pre-trained model.}
    \label{tab:rt_embedders_long}
    \centering
    \begin{tabular}{ll *{40}c}
        \toprule
        Dataset & Embedding & Mean & StD & Valid & Cached \\
        \midrule
         & Concat & $0$ & $0$ & 4543 & 0  \\
        Appliances energy & Catch22 & $99$ & $26$ & 7442 & 0  \\
        \vspace{1mm}
         & TS2Vec & $327$ & $125$ & 40 & 22110  \\
        \rowcolor{gray!25}
         & Concat & $0$ & $0$ & 9099 & 0  \\
        \rowcolor{gray!25}
        ElectricDevices & Catch22 & $2$ & $2$ & 9855 & 0  \\
        \rowcolor{gray!25}
        \vspace{1mm}
         & TS2Vec & $391$ & $107$ & 40 & 27368  \\
         & Concat & $0$ & $0$ & 4574 & 0  \\
        Exchange rate & Catch22 & $3$ & $2$ & 7859 & 0  \\
        \vspace{1mm}
         & TS2Vec & $41$ & $16$ & 41 & 22220  \\
        \rowcolor{gray!25}
         & Concat & $0$ & $0$ & 4193 & 0  \\
        \rowcolor{gray!25}
        Google stock & Catch22 & $1$ & $2$ & 7221 & 0  \\
        \rowcolor{gray!25}
        \vspace{1mm}
         & TS2Vec & $38$ & $20$ & 40 & 22154  \\ 
         & Concat & $0$ & $0$ & 5537 & 0  \\
        PPG and respiration & Catch22 & $14$ & $4$ & 7817 & 0  \\
        \vspace{1mm}
         & TS2Vec & $515$ & $259$ & 40 & 22286  \\
        \rowcolor{gray!25}
         & Concat & $0$ & $0$ & 11700 & 0  \\
        \rowcolor{gray!25}
        PTB diagnostic ECG & Catch22 & $889$ & $162$ & 10394 & 0   \\
        \rowcolor{gray!25}
        \vspace{1mm}
         & TS2Vec & $3697$ & $1035$ & 39 & 26884  \\
         & Concat & $0$ & $0$ & 10071 & 0  \\
        Sine & Catch22 & $3$ & $2$ & 10682 & 0 \\
        \vspace{1mm}
         & TS2Vec & $252$ & $94$ & 40 & 29920 \\ 
        \rowcolor{gray!25}
         & Concat & $0$ & $0$ & 9727 & 0 \\
        \rowcolor{gray!25}
        StarLightCurves & Catch22 & $12$ & $6$ & 10622 & 0  \\
        \rowcolor{gray!25}
        \vspace{1mm}
         & TS2Vec & $640$ & $364$ & 40 & 27269  \\
         & Concat & $0$ & $0$ & 5717 & 0 \\
        UniMiB SHAR & Catch22 & $7$ & $3$ & 10633 & 0  \\
        \vspace{1mm}
         & TS2Vec & $256$ & $113$ & 41 & 29909 \\
        \rowcolor{gray!25}
         & Concat & $0$ & $0$ & 7509 & 0 \\
        \rowcolor{gray!25}
        Wikipedia web traffic & Catch22 & $86$ & $23$ & 5182 & 0 \\
        \rowcolor{gray!25}
         & TS2Vec & $4812$ & $1475$ & 36 & 17149 \\
        \bottomrule
    \end{tabular}
\end{table}

\section{Measure Selection Guide}
\label{sec:appendix:guide}
Ultimately, STEB analyzes and compares synthetic TS evaluation measures to allow users a better selection of measures. Below, we provide some step-by-step instructions on how this selection process can look like using STEB output.
\begin{enumerate}
    \item Determine which categories are relevant for the given use case. Usually, one measure is needed per category. For each category, repeat all following steps.
    \item Start with the measure ranked highest reliability (see \cref{tab:ranking}).
    \item If the measure lacks consistency in this category (see \cref{tab:consistency}), move on to the next best measure. We suggest a minimum of $r_\text{con} = 0.5$.
    \item If the use case dictates running time constraints, for instance, because many parametrizations of a generative model must be evaluated during optimization, measures with excessive running time should be skipped (see \cref{tab:rt_measures} in combination with \cref{tab:rt_embedders}). For this purpose, the running time on the dataset most similar in size, TS length, and number of feature channels to the use case's target dataset should be considered.
    \item Further, in case ease-of-use is an important selection criterion, it makes sense to skip embedder-dependent measures or only use them in case embeddings can be reused in other categories. Similarly, some measures are prone to errors such as DOMIAS and $C_T$, which requires additional effort and knowledge to fix (see \cref{tab:statistics} and \cref{tab:failures}).
\end{enumerate}

Based on the experimental results in this paper and assuming running time is no limiting factor, the best measure suit for synthetic TS evaluation currently comprises
\begin{itemize}
    \item \emph{\alphaprecision} for fidelity,
    \item \emph{\acs} for generalization,
    \item \emph{\autocorr} for privacy, and
    \item \emph{\contextfid} for representativeness.
\end{itemize}

\end{document}